\documentclass[11pt]{article}

\usepackage{acl}

\usepackage{latexsym}

\usepackage[T1]{fontenc}
\usepackage{lmodern}
\usepackage{textcomp}
\usepackage{times}
\usepackage[utf8]{inputenc}

\usepackage{microtype}

\usepackage{inconsolata}

\usepackage{graphicx}
\usepackage{xspace}
\usepackage[textsize=tiny]{todonotes}
\usepackage{cleveref}
\usepackage{enumitem}
\usepackage{booktabs}
\usepackage{multirow}
\usepackage{adjustbox}
\usepackage{mdframed}
\usepackage{array}
\usepackage{amsthm}
\theoremstyle{definition} 
\usepackage{float}
\usepackage{etoolbox}

\newmdtheoremenv[%
linecolor=gray,leftmargin=10,%
rightmargin=10,
backgroundcolor=gray!40,
nobreak=false,
]{myprop}{Template}[section]

\usepackage{listings}
\definecolor{mypurple}{RGB}{243, 243, 255}
\definecolor{mygreen}{RGB}{228, 242, 228}
\definecolor{myred}{RGB}{233, 217, 217}

\newcommand{\methodname}{\textsc{unspecific}\xspace}
\newcommand{\heading}[1]{\vspace{5pt}\noindent\underline{\textsc{#1}}}
\newcommand{\headingbase}[1]{\noindent\underline{\textsc{#1}}}
\newcommand{\headingexp}[1]{\noindent\textbf{#1}}

\title{\methodname: General Constraint Synthesis for Breaking Copy-and-Paste Shortcut in LLM Instruction Following}

\author{
Jeet Sharma\thanks{\, indicates equal contribution.}\textsuperscript{1} \;\; Balpreet Kaur\footnotemark[1]\textsuperscript{1} \;\; Jeremiah Hong\textsuperscript{1} \\ 
\textbf{Hamed Zamani\textsuperscript{1}} \;\; \textbf{Haw-Shiuan Chang\footnotemark[1]\textsuperscript{1}} \\
        \textsuperscript{1}University of Massachusetts, Amherst, USA \\
\texttt{ \{jeetdevendra,bbalpreetkau,jshong\}@umass.edu}, \\ \texttt{ \{zamani,hschang\}@cs.umass.edu}
}

\begin{document}
\maketitle
\begin{abstract}

Large language models (LLMs) are increasingly expected to follow long lists of constraints in complex instructions, and synthesizing instructions from a reference document  (i.e., back-translation) is a widely used method to measure/enhance LLMs’ ability to follow complex instructions. However, this method introduces a critical loophole: the constraint synthesis model copies text from the reference as a very specific constraint and the evaluated LLM trivially satisfies the constraint by copying its text in the response. 
To address these issues, we propose \methodname, a novel framework that synthesizes constraints common to two similar reference articles to reduce copy-pasting, selectively hardens only trivially satisfied constraints to balance difficulty and naturalness, and evaluates satisfaction on both the generated article and its summary to penalize superficial instruction following. Consequently, we built the \methodname benchmark on news, story, and blog domains to analyze the copy-pasting behavior of LLMs. Our results show that our synthesized constraints are not only more challenging (e.g., the satisfaction rate of \textit{GPT-5 Mini} drops from $90$\% to $78$\%) and natural (LLM win-rate gap improves by $30$\%) from a human perspective but also mitigate the copy-pasting. We also find that a large portion of constraints are satisfied superficially (i.e., not satisfied in the core narrative of the article). The code and datasets are released at 
\url{https://github.com/JeetDSharma/UNSPECIFIC}.
%\url{https://anonymous.4open.science/r/UNSPECIFIC-4DC6}.

% 

% propose a benchmark

% news, story, blog

%Experiments demonstrate that \methodname produces harder, more natural benchmarks, exposes significant variation in LLMs' true instruction-following ability, and quantifies copy-paste reliance across model families and domains.

% %improve automatic instruction following benchmark construction

\end{abstract}

\section{Introduction}

When users interact with a large language models (LLM), whether the LLM could faithfully follow instruction often directly determines users' perception of the LLMs' ability. Nowadays, LLM agents often need to understand and respect many constraints in a long context or multi-turn dialog for generating an article, so the ability of understanding and thoroughly satisfying a long list of constraints or context is crucial for excellent user experiences. 
%For LLMs, the success rates of following instructions usually degrade as the instruction contains more and more constraints~\citep{sun2024conifer,atmakuru2024cs4,lu2025benchmarking,jaroslawicz2025many,zhang2025cfbench,guo2025recast}, 
Sometimes, the LLMs seem to follow all the instructions, but a closer examination shows that they satisfy them in a superficial way without understanding the meaning of the constraints. For example, given a constraint \textit{``the story should have a happy ending''}, the LLM might just abruptly append \textit{``this story has a happy ending''} at the end.

\begin{figure}[t]
  \includegraphics[width=\columnwidth]{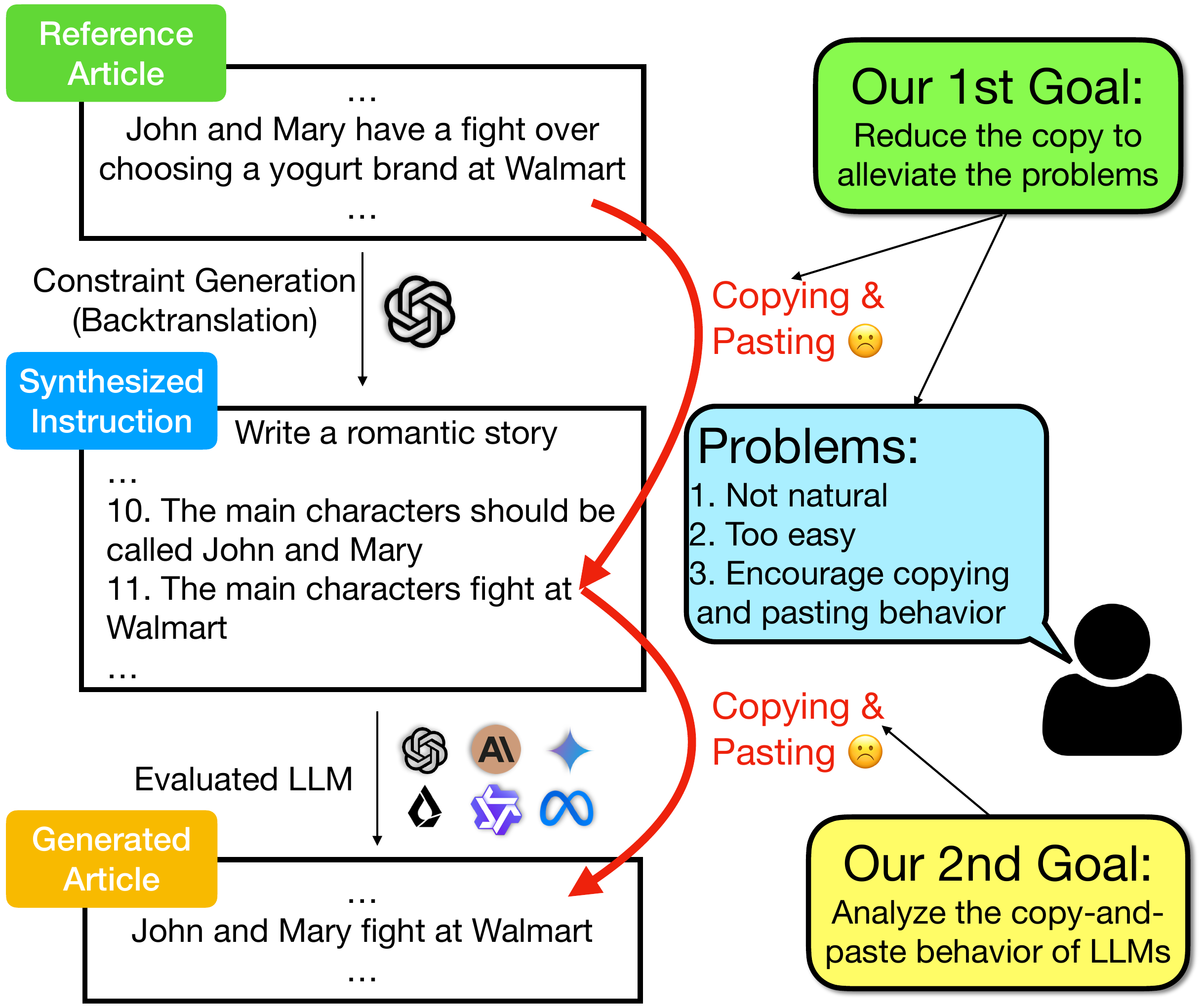}
  \caption{The existence of the copy-and-paste shortcuts make some constraints synthesized from the reference article too specific to be natural from the real users and too easy to accurately measure instruction-following ability of recent LLMs. Our goals are discouraging the LLM from copying reference article into constraints and analyzing LLM's copying behavior when generating articles.}
  \label{fig:problem}
\end{figure}

As the number of constraints increases in a complex instruction, the problem of constraint violation or superficial constraints following becomes more serious~\citep{sun2024conifer,atmakuru2024cs4,lu2025benchmarking,jaroslawicz2025many,zhang2025cfbench,guo2025recast,ye2026muldimif}, so researchers need a large amount of instructions with many challenging and natural constraints to measure or improve the instruction following ability of the state-of-the-art LLMs. However, hiring experts to write the constraints is very expensive and hard to scale~\citep{wang2023self,xu2024wizardlm}.

Synthesizing constraints by LLMs is a widely-adopted alternative. A common approach is to ask an LLM to extract constraints from a reference article, which can ensure there exists a response that satisfies all the constraints~\citep{pham-etal-2024-suri, atmakuru2024cs4}. The method is sometimes called back-translation~\citep{li2024self, nguyen2024better, chen2024dog, qi2025constraint}. However, the reference-based constraint generation approach has a key weakness illustrated in \Cref{fig:problem}: the constraint synthesis LLM sometimes simply copies and pastes from the reference article and the evaluated LLMs could simply copy and paste the constraint into its generated article.

The loophole causes three problems: 
First, the synthesized constraints are often too specific to be natural because the users typically do not have a clear idea about the details in the article they want to write. 
Second, too much information of the reference article is leaked through the constraints, which makes the constraints in the benchmark very easy to be satisfied. Moreover, LLMs could often just reorder the constraints and add some transitions to reconstruct the original reference article. 
Third, if LLMs are trained to optimize the satisfaction rate of the synthesized constraints, the LLMs are encouraged to take the copy-and-paste shortcuts. This might explain why we often observe the superficial instruction following behavior from the current LLMs.
%satisfy the constraints superficially by simply copying and pasting without understanding users' intention. 

To address the issues, we propose a novel constraint synthesis and evaluation framework: \methodname (\textbf{UN}covering \textbf{S}ummary-resistant, \textbf{P}aste-reducing, and \textbf{E}valuation-revised \textbf{C}onstraints to \textbf{I}mprove \textbf{F}ollowing \textbf{I}nstruction \textbf{C}hallenge). We first ask an LLM to generate the common constraints from two similar reference articles. This could prevent the LLM from including the specific details that only exist in one of the articles. Next, we ask the LLM to revise the constraints that are coincidentally satisfied by an LLM-generated article. Finally, we evaluate the constraint satisfaction not only on the original generated article but also on the summary of the article. To achieve a high satisfaction rate after the summarization, the evaluated LLM needs to satisfy the constraints in the core narrative rather than in details.

In our experiments, we show that \methodname increases the difficulty of the instruction following tasks without sacrificing their naturalness, reduces LLMs' copying behavior, and provides metrics to quantify the behavior. We synthesize \methodname benchmark that measures how often LLMs could deeply integrate the constraints into the article when encountering the general constraints and specific constraints in news, story, and blog domains. The results show that two LLMs could perform similarly given specific constraints synthesized from a single reference article while satisfying very different percentages of general common constraints from two similar reference articles. We also discover that some LLMs tend to achieve much better performance by relying on constraint copying.

\begin{figure*}[t]
  \includegraphics[width=1\linewidth]{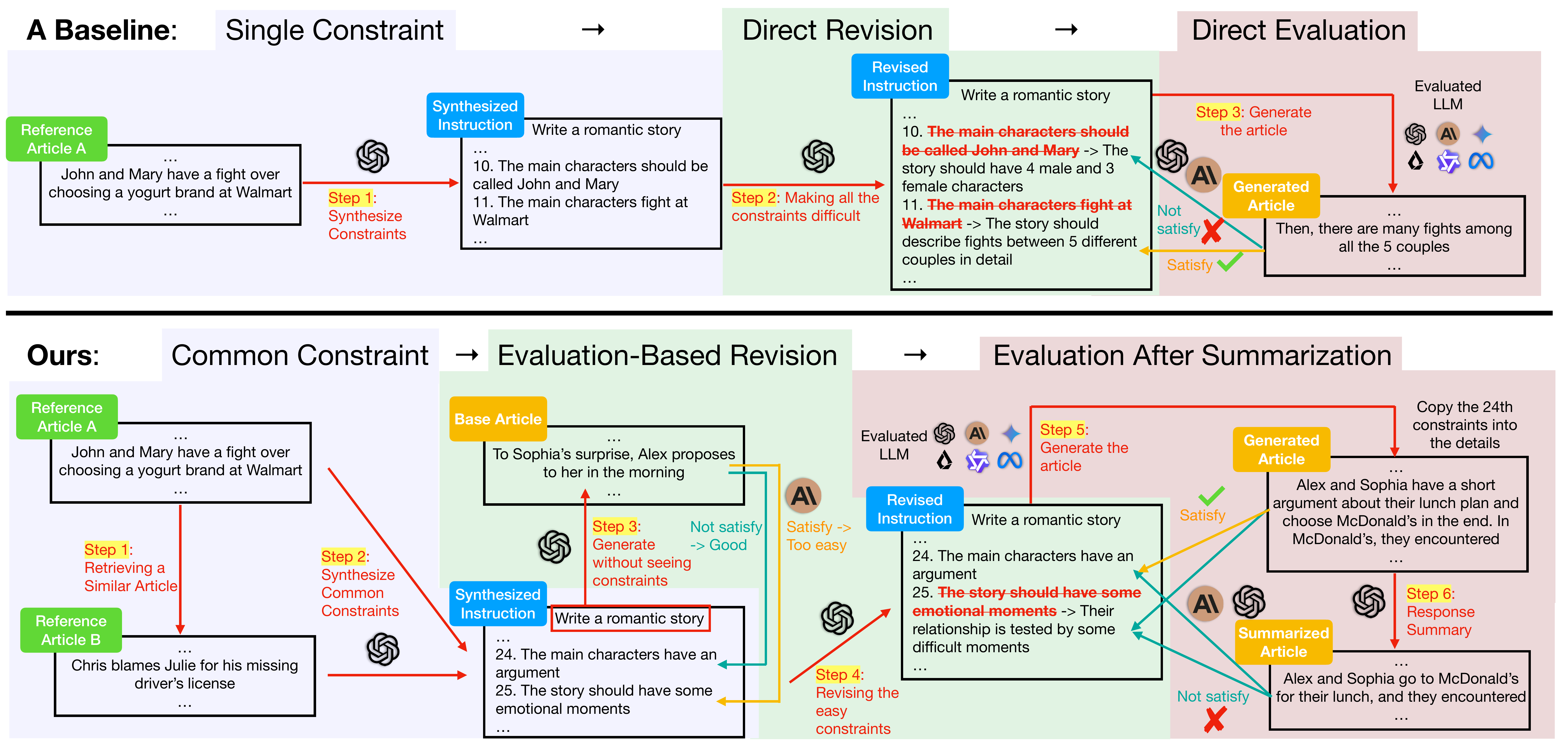}
  \caption{\fboxsep=1pt Comparison of our three main novel components in \methodname at bottom with their corresponding baselines on top. \colorbox{mypurple}{\textit{Left}}: Our method synthesizes more general constraints that are satisfied by both input articles simultaneously. 
  \colorbox{mygreen}{\textit{Middle}}: Our method first generates a base article just from the main task and then increases only the difficulties of the constraints satisfied by this base article. \colorbox{myred}{ \textit{Right}}: Our method summarizes the generated article to remove the details and check how many constraints are still satisfied by the summary, which mostly only contains the core narrative of the generated article.
  }
  \label{fig:workflow}
\end{figure*}

%verify 
%that 
%our approach could 
%synthesize the constraints that mitigate the copy and paste behavior and encourage the LLMs to incorporate the constraints in the main narrative flow.
%balance the naturalness and difficulty
%human experiments

%general constraints, specific constraints
%satisfying the constraints in a general way and specific way

%analyses
%Some LLMs copy and paste more.
%Some LLMs handle the abstract constraints better

\heading{Main Contributions}
\begin{itemize}[itemsep=2pt, parsep=0pt, topsep=1pt,leftmargin=*]
    \item Identify the copying shortcut problem in reference-based constraint synthesis and instruction following.
    \item We propose to synthesize the common constraints from two similar articles. Compared to synthesizing the constraints from a single article, our approach generates the more general constraints and reduces LLMs' copy-and-paste behavior. 
    \item We propose the evaluation-based constraint revision that increases the difficulties of only easy constraints. The approach generates more natural instructions compared to revising all the constraints.
    \item We propose to evaluate the constraint satisfaction rate after the summarization. The new metric measures the percentage of the constraints that are not satisfied superficially in the generated article. 
\end{itemize}

\section{Method}

To simplify our experiments, we assume that all the instructions have a main task and $39$ constraints as in \citet{atmakuru2024cs4}. Our goal is to synthesize natural and challenging instructions from a corpus and the instructions cannot be easily satisfied by copy and paste. 

The copying behavior reduces the difficulty of instruction following and one intuitive solution is to ask the LLM to make the constraints harder. The top row of \Cref{fig:workflow} illustrates this baseline. However, the LLM usually completes the task by adding lots of details into the constraints, which makes the constraints unnatural and sometimes encourages the evaluated LLM to copy even more details. 

To increase the difficulty without sacrificing naturalness or inducing more copying, we propose to generate the common constraints and revise the constraints based on the evaluation results. Additionally, we propose a new evaluation method that could quantify the copying behavior. The bottom row of \Cref{fig:workflow} summarizes our methods.

%how to prevent constraints to be too general
%some general constraints are too easy
%so revision
%how to evaluate copy and paste

%this approach could make the instruction 

%goal
%main task and constraints
%39 constraints~\citep{atmakuru2024cs4}
%create the instructions that are more general but still challenging

%the constraints would be more natural, challenging, cannot be easily satisfied by copy and paste
%able to quantify copy and paste behavior
%too easy -> make the constraints more difficult

%describe baseline
%copy and paste

\subsection{Common Constraints from Two Articles}
\label{sec:common}

The users often do not have a very concrete idea about what they want to write. Otherwise, they will write (the outline of) the article by themselves. Thus, \textbf{if one synthesized constraint could only be satisfied by an existing article, the constraint is often too specific and too rare to be natural}. 

To avoid generating such constraints from only the single reference document A, we first retrieve another similar document B using sentence BERT~\citep{reimers2019sentence} and ask the LLM to generate the common constraints, all of which are satisfied by article A and by article B simultaneously. See an example in \Cref{fig:workflow}. The procedure ensures that the synthesized constraint set could be satisfied by multiple articles and reduce the information leaking of individual article. Notice that the lower similarity between A and B leads to more general constraints, so we can control the generality of the constraints by choosing the article B. In our benchmark, we retrieve the article that has the highest similarity less than $0.85$ to exclude the near duplicated articles. %\todo{confirm this}.

%The constraint is too specific if it only applicable to one article.
%too specific
%too rare to be natural
%from the two similar articles

%We can control the similarity of the retrieved article to 
%0.7

%retrieve similar documents
%need to be the constraints of the article A and article B simultaneously

%A natural constraint should at least appear in 2 stories
%The whole set of constraints could lead to multiple stories

\subsection{Evaluation-based Constraint Revision}
\label{sec:eval_rev}

Synthesizing general constraints might lead to an undesirable side effect: the resulting constraints (e.g., \textit{The story should have a strong ending}) might be too general to be challenging. Directly revising all constraints could trade off difficulty against the naturalness, but our goal is to achieve both criteria at the same time. 

To break the tradeoff, we ask LLMs to only revise the general constraints that are trivially satisfiable. An informative constraint should rarely be satisfied without being specified. To identify the redundant constraints, we first ask \textit{GPT-4.1 Mini}
%~\todo{to be confirm} 
to generate a base article just according to only the main task without showing any constraints. Next, we ask \textit{GPT-4.1 Mini} to replace the constraints that are satisfied by this base article with harder ones as shown in \Cref{fig:workflow}. In this way, we can change fewer constraints than the typical adversarial method~\citep{zellers2019hellaswag}, which would provide the full instruction including constraints, and preserve the high naturalness. Nevertheless, we surprisingly find that around $50$-$60$\% of the common constraints from two similar articles and around $20$-$30$\% of the constraints from a single article are so easy that LLMs could satisfy them without even seeing them in the input. See more details in \Cref{tb:revision_perc}.
%are satisfied by this base article and thus need to be revised, %that partially explain why the current LLMs could satisfy most of the synthesized constraints. 

%Notice that this method is much more conservative than 
%different from 
%we did not provide the full input

%we conduct experiment
%percentage of satisfying the constraints (single and common constraints)

%increase the difficulty at the cost of naturalness. 
%One intuitive method is to ask LLM to directly revise all constraints to increase the difficulty.
%informative, which means almost all the articles LLM generates would satisfy the constraints. 
%tradeoff between naturalness and difficulty
%direct revision

%Making constraints general 
%One concern 
%is so general
%almost all the articles LLM generates would satisfy the constraints
%too general to be informative
%dilemma

\subsection{Evaluation After Summarization}
\label{sec:core_eval}
After synthesizing many constraints that encourage the evaluated LLMs to deeply integrate the constraints into their core narrative, we need a metric to test how much the LLMs could achieve that. The text similarity or overlap detection between the constraints and the response might not provide reliable metrics because the evaluated LLMs could simply paraphrase the constraints~\citep{ippolito2023preventing} to achieve a higher score.

To examine whether the LLMs follow the constraints using its core narrative, we use \textit{GPT-4.1 Mini} to summarize the response before evaluating the constraint satisfaction rate. If the LLM only satisfies the constraints superficially using the details of its response, the LLM cannot achieve high satisfaction rate after the summarization removes these details. In our experiments, we control the summary to have 25\% of its original lengths and this percentage could be tuned to measure how deep the constraints are integrated into the core narrative.

%as the article with only 25\% of its original length
%constraints would be removed after and 
%put the constraints in the details of response
%how 

%summary without seeing the constraints
%the superficial constraint satisfaction tends to be removed
%main narrative
%details

\begin{table*}[t]
    \centering
    \scalebox{0.9}{
    \begin{tabular}{cc|cccc|cccc|ccc}
        \toprule
        & & \multicolumn{4}{c|}{LLM Satisfaction Judge} & \multicolumn{4}{c|}{LLM Coherence Judge} & \multicolumn{3}{c}{Human Naturalness Judge} \\
       \midrule
        & & \multicolumn{2}{c}{GPT-5 Mini} & \multicolumn{2}{c|}{Llama-3 8B} & \multicolumn{2}{c}{GPT-5 Mini} & \multicolumn{2}{c|}{Llama-3 8B} & Human & \multirow{2}{*}{Tie} & LLM \\
        Synth & Revise & Org & Core & Org & Core & Org & Core & Org & Core   & Win &  & Win   \\
        \midrule
        Single & None & 89.7 & 48.4 & 50.7 & 15.2 & 3.36 & 2.80 & 2.32 & 2.17 & 59 & 5 & 36 \\
        Common & None & 95.1 & 65.7 & 57.9 & 27.7 & 3.52 & 2.92 & 2.44 & 2.36 & 40 & 14 & 46 \\
        Common & Direct & 71.8 & 41.0 & 27.6 & 8.2 & 3.22 & 2.72 & 2.36 & 2.16 & 48 & 3 & 49 \\
        Common & Eval & 78.2 & 46.5 & 38.1 & 10.6 & 3.28 & 2.88 & 2.28 & 2.12 & 41 & 11 & 48 \\
        \bottomrule
    \end{tabular}
    }
    \caption{Comparison of different constraint synthesis methods in news domain. Single+None is the back-translation baseline and Common+Eval is our method. The numbers except for coherence are percentages.}
    \label{tb:human_exp}
\end{table*}

\begin{table*}
    \centering
    \begin{tabular}{p{7cm}|p{7cm}}
    \toprule
        Single Constraints & Common Constaints \\
    \midrule
        1. Begin by describing high schoolers walking out of class across Alaska & 1. Begin with a scene that involves students participating in a protest\\
        2. Include details about students chanting slogans and marching to the state Capitol & 2. Include detailed accounts of the middle portion of events \\
        3. Conclude by emphasizing student activism as democracy in action & 3. End with a statement reflecting on the broader significance of student activism \\
    \bottomrule
    \end{tabular}
    \caption{Example comparison between the constraints from a single reference article and the common constraints from two similar reference articles}
    \label{tb:common_examples}
\end{table*}

% \begin{table}[t]
%     \centering
%     \scalebox{1}{
%     \begin{tabular}{cc|ccc}
%         \toprule
%         & & Human & \multirow{2}{*}{Tie} & LLM \\
%         Synth & Revise & Win &  & Win   \\
%         \midrule
%         Single & None & \\
%         Common & None & \\
%         Common & Direct & \\
%         Common & Eval & \\
%         \bottomrule
%     \end{tabular}
%     }
%     \caption{News}
%     \label{tb:human_exp}
% \end{table}

% \begin{table}[t]
%     \centering
%     \scalebox{0.8}{
%     \begin{tabular}{ccc|cccc}
%         \toprule
%          &  &  & \multicolumn{2}{c}{GPT-5 Mini} & \multicolumn{2}{c}{Llama-3 8B} \\
%         Task & Synth & Revise & Org & Core & Org & Core \\
%          \midrule
%         \multirow{9}{*}{Gen} & \multirow{3}{*}{Single} & None &  89.7 & 48.4 & 50.7 & 15.2 \\
%         & & Direct & \\
%         & & Eval & \\
%         \cmidrule(lr){2-7} 
%          & General & None &   \\
%         & Constraint & Direct & \\
%         & Prompting & Eval & \\
%         \cmidrule(lr){2-7} 
%          & \multirow{3}{*}{Common} & None &   \\
%         & & Direct & \\
%         & & Eval & \\
%         \midrule
%         \multirow{6}{*}{Edit} & \multirow{3}{*}{Single} & None &   \\
%         & & Direct & \\
%         & & Eval & \\
%         \cmidrule(lr){2-7}
%          & \multirow{3}{*}{Common} & None &   \\
%         & & Direct & \\
%         & & Eval & \\
%          \bottomrule
%     \end{tabular}
%     }
%     \caption{The satisfaction rate in the news domain}
%     \label{tb:baselines_sat}
% \end{table}
\begin{table}[t]
    \centering
    \scalebox{0.8}{
    \begin{tabular}{ccc|cccc}
        \toprule
         &  &  & \multicolumn{2}{c}{GPT-5 Mini} & \multicolumn{2}{c}{Llama-3 8B} \\
        Task & Synth & Revise & Org & Core & Org & Core \\
         \midrule
        \multirow{9}{*}{Gen} & \multirow{3}{*}{Single} & None &  89.7 & 48.4 & 50.7 & 15.2 \\
        & & Direct & 70.2 & 31.1 & 33.3 & 7.5 \\
        & & Eval & 77.5 & 35.5 & 38.2 & 9.4 \\
        \cmidrule(lr){2-7} 
         & General & None & 90.4 & 47.6 & 53.3 & 31.6 \\
        & Constraint & Direct & 70.1 & 28.1 & 31.7 & 13.1 \\
        & Prompting & Eval & 79.8 & 28.1 & 32.5 & 12.9 \\
        \cmidrule(lr){2-7} 
         & \multirow{3}{*}{Common} & None & 95.1 & 65.7 & 57.9 & 27.7 \\
        & & Direct & 71.8 & 41.0 & 27.6 & 8.2 \\
        & & Eval & 78.2 & 46.5 & 38.1 & 10.6 \\
        \midrule
        \multirow{6}{*}{Edit} & \multirow{3}{*}{Single} & None & 86.6 & 41.9 & 30.1 & 8.2 \\
        & & Direct & 70.5 & 27.3 & 14.7 & 3.2 \\
        & & Eval & 77.1 & 23.7 & 25.1 & 5.9 \\
        \cmidrule(lr){2-7}
         & \multirow{3}{*}{Common} & None & 93.3 & 62.7 & 52.1 & 23.7 \\
        & & Direct & 72.8 & 36.6 & 21.0 & 5.2 \\
        & & Eval & 77.2 & 44.6 & 34.8 & 6.5 \\
         \bottomrule
    \end{tabular}
    }
    \caption{The satisfaction rate in the news domain. The Org is the satisfaction rate for the original response and Core is the satisfaction rate for the summary of the response with $25$\% length of the response. The maximum standard error across all cells is 4.78. }
    \label{tb:baselines_sat}
\end{table}
% \begin{table}[t]
%     \centering
%     \scalebox{0.8}{
%     \begin{tabular}{ccc|cccc}
%         \toprule
%          &  &  & \multicolumn{2}{c}{GPT-5 Mini} & \multicolumn{2}{c}{Llama-3 8B} \\
%         Task & Synth & Revise & Org & Core & Org & Core \\
%          \midrule
%         \multirow{9}{*}{Gen} & \multirow{3}{*}{Single} & None &  0.589 & 0.602 & 0.501 & 0.508 \\
%         & & Direct & \\
%         & & Eval & \\
%         \cmidrule(lr){2-7}
%         & General & None &   \\
%         & Constraint & Direct & \\
%         & Prompting & Eval & \\
%         \cmidrule(lr){2-7} 
%          & \multirow{3}{*}{Common} & None &   \\
%         & & Direct & \\
%         & & Eval & \\
%         \midrule
%         \multirow{6}{*}{Edit} & \multirow{3}{*}{Single} & None &   \\
%         & & Direct & \\
%         & & Eval & \\
%         \cmidrule(lr){2-7}
%          & \multirow{3}{*}{Common} & None &   \\
%         & & Direct & \\
%         & & Eval & \\
%          \bottomrule
%     \end{tabular}
%     }
%     \caption{The average similarity between each constraint to the closest sentence in the response.}
%     \label{tb:baselines_sim}
% \end{table}

\begin{table}[t]
    \centering
    \scalebox{1}{
    \begin{tabular}{c|ccc}
        \toprule
        & Single & General & Common \\
        \midrule
        None & 0.546 & 0.514 & 0.487 \\
        Direct & 0.509 & 0.498 & 0.463 \\
        Eval & 0.540 & 0.512 & 0.489 \\
        \bottomrule
    \end{tabular}
    }
    \caption{The average similarity between each constraint to the closest sentence in the reference article(s). The maximum Standard error across all cells is 0.018.}
    \label{tb:ref_sim}
\end{table}
\begin{table}[t]
    \centering
    \scalebox{0.8}{
    \begin{tabular}{ccc|cccc}
        \toprule
         &  &  & \multicolumn{2}{c}{GPT-5 Mini} & \multicolumn{2}{c}{Llama-3 8B} \\
        Task & Synth & Revise & Org & Core & Org & Core \\
         \midrule
        \multirow{9}{*}{Gen} & \multirow{3}{*}{Single} & None &  0.589 & 0.501 & 0.602 & 0.508 \\
        & & Direct & 0.597 & 0.528 & 0.638 & 0.523 \\
        & & Eval & 0.598 & 0.509 & 0.617 & 0.512 \\
        \cmidrule(lr){2-7}
        & General & None & 0.581 & 0.500 & 0.599 & 0.527 \\
        & Constraint & Direct & 0.594 & 0.544 & 0.640 & 0.558 \\
        & Prompting & Eval & 0.579 & 0.510 & 0.615 & 0.542 \\
        \cmidrule(lr){2-7} 
         & \multirow{3}{*}{Common} & None & 0.546 & 0.459 & 0.558 & 0.482 \\
        & & Direct & 0.579 & 0.498 & 0.614 & 0.512 \\
        & & Eval & 0.561 & 0.485 & 0.608 & 0.509 \\
        \midrule
        \multirow{6}{*}{Edit} & \multirow{3}{*}{Single} & None & 0.586 & 0.501 & 0.565 & 0.485 \\
        & & Direct & 0.604 & 0.525 & 0.595 & 0.509 \\
        & & Eval & 0.596 & 0.518 & 0.603 & 0.511 \\
        \cmidrule(lr){2-7}
         & \multirow{3}{*}{Common} & None & 0.541 & 0.449 & 0.534 & 0.450 \\
        & & Direct & 0.577 & 0.497 & 0.607 & 0.497 \\
        & & Eval & 0.570 & 0.490 & 0.590 & 0.479 \\
         \bottomrule
    \end{tabular}
    }
    \caption{The average similarity between each constraint to the closest sentence in the response. Maximum standard error across all cells is 0.024.}
    \label{tb:baselines_sim}
\end{table}

\section{Constraint Synthesizing Experiments}
In this section, we compare our proposed constraint synthesis and evaluation methods with strong baselines to verify their effectiveness. %analyze LLMs' copy-and-paste behavior using these methods. 

\subsection{Setup}
%news\todo{Balpreet: explain how we prepare news data}

We obtained news articles from 3DLNews2 \citep{ariyarathne_nwala_3dlnews}, a large-scale dataset of US local news articles spanning nearly three decades (1995–2024). The news is collected from over $14,000$ local newspapers, TV stations, and radio broadcasters across all $50$ states. We choose local news to reduce the chance that LLMs remember the contents of the well-known reference news. %avoid evaluated LLMs from directly outputting the well-known news events based on its memory.
For each method, we synthesize $25$ instructions and each instruction has $39$ constraints, so there are $975$ constraints in total, which we find provide stable experiment results to support our conclusion. 
%To reduce the evaluation variance, one of the reference article in \underline{Common} is the same  \underline{Single}
%\todo{mention this. We find that this number of examples could already give us enough signal to analyze the copy behavior}

All the constraint synthesis methods using \textit{GPT-4.1 Mini}. To prevent the order of the constraints from having the same order of relevant sentences in the reference story, we instruct the LLM to shuffle the constraints. %The LLM synthesizes $39$ constraints for each main task
%each constraint satisfaction rate is computed by the average of . 
%975 constraint satisfaction evaluation
%shuffle the constraint order ()
Next, we use \textit{GPT-4.1 Mini} to generates the base articles with around $500$ words from just the main tasks without seeing the constraints. The base articles are utilized in our evaluation-based revision as described in \Cref{sec:eval_rev}, editing experiments, and article quality evaluation. 
%Besides, we also test 

To test the difficulty of the constraints and analyze the copy-and-paste behavior, we select two representative LLMs, \textit{GPT-5 Mini} and \textit{Llama-3 8B}, to generate the articles with around $500$ words that satisfy the constraints. Besides generating from scratch, we ask the two LLMs to edit the base articles to satisfy the constraints. 

%choose \textit{GPT-5 Mini} as a strong LLM and \textit{Llama-3 8B} as a weak LLM.
%generate $500$ words
%Generation vs editing
%The articles are generated by \textit{GPT-5 Mini} and \textit{Llama-3 8B}. 

\subsubsection{Baselines}

The following synthesizing methods are tested:

\headingbase{Single:} The standard back-translation method using a single reference article.

\headingbase{Common:} Starting from the same reference article in the single baseline, we first retrieve a similar article using \texttt{all-mpnet-base-v2} and generate the common constraints from the two articles (\Cref{sec:common}).

\headingbase{General constraint prompting:} We ask the LLM to synthesize general constraints from a single reference article.

The following revision methods are tested:
\headingbase{None:} No revision.

\headingbase{Direct:} Ask the LLM to increase the difficulties of all constraints.

\headingbase{Eval:} Revise only the trivially satisfied constraints (\Cref{sec:eval_rev}).

\subsubsection{Automatic Article Evaluation}

We evaluate the generated article using its satisfaction rates, quality, and similarity to the constraints. The quality and constraint satisfaction rates of articles are evaluated by \textit{Claude Sonnet 4}. 
For each constraint satisfaction, the LLM judge outputs yes or no followed by a brief explanation. The average yes ratio is called original satisfaction rate (\underline{Org}). After summarizing the article into 25\% of its original length, the metric described in \Cref{sec:core_eval} is called core satisfaction rate (\underline{Core}). 

The evaluated LLM might pursue a high satisfaction rate while sacrificing the quality by piecing all the constraints into a non-coherent article. To understand the relevancy-quality trade-off of different LLMs, we also use the LLM judge to evaluate the coherence and likability. To avoid the over-generous score problem from some LLMs, each article is compared with the base article, which is generated by \textit{GPT-4.1 Mini} with only the main task input.

To further analyze the copy-and-paste behavior, we use sentence BERT \texttt{all-mpnet-base-v2} to retrieve the most similar sentence in the response or reference article to each constraint and average the similarity across all the constraints. The higher similarity suggests the LLM copies and pastes more.

%\todo{The baseline of our core satisfactory rate metric? compute the similarity with the reference article?}
%LLM judge using Sonnet 
%with explanation

%satisfaction
%summarization
%core and org

%\todo{when we evaluate the quality, do we also output explanation?} 

%satisfying many constraints might sacrificing the coherence.
%Coherence, compare with the base article and report the average score. 
%from \textit{GPT-4.1 Mini} 
%likeness 
%generous

%Similarity metrics
%

\subsubsection{Naturalness Human Evaluation}

It is not practical to evaluate LLMs on the synthesized difficult constraints that users are unlikely to specify, so the naturalness of the constraints is also an important metric for benchmark construction. 
%It is easy to synthsize very complex constraints.
%evaluate the constraints

% Use MTurk to collect human constraints

% 25 rows, two authors of the paper
%\todo{the human experiment setup}
% inter-annotator agreement
% agreement 60.0\%
% Cohen's $\kappa$ 0.310
%We assess naturalness through a crowdsourced annotation study on Amazon Mechanical Turk (MTurk). 

In our human experiments, we first ask workers from Amazon Mechanical Turk (MTurk) to write the constraints relevant to our news writing main tasks. Next, we ask two experts, who have a computer science master degree, to compare the naturalness of human-written constraints and the LLM-synthesized constraints. 
%\textbf{Inter-annotator agreement.} 
Each annotator independently annotates $100$ comparisons ($25$ instructions across $4$ methods); $60$\% of their judgments are the same, which is much larger than the $33$\% expected by random guessing, and the Cohen's $\kappa$ is $0.310$.

%The central element of every task is a writing prompt, a specific news article topic, paired with a constraint, an instruction a writer might follow when responding to that prompt. Annotators judged how natural each constraint felt for its corresponding prompt.

%LLM as a judge?

%\subsection{Common Constraint Validation}
\subsection{Constraint Synthesis Comparison}

We evaluate the proposed common constraint synthesis (\underline{Common}) and evaluation-based revision (\underline{Eval}) below: 

\headingexp{\underline{Common+Eval} is more challenging and natural} compared to \underline{Single+None} baseline in \Cref{tb:human_exp}. The original satisfaction rate drops from $89.7$ to $78.2$ for \textit{GPT-5 Mini} and from $50.7$ to $38.1$ for \textit{Llama-3 8B}, the coherence remains similar, and the naturalness winning rate difference compared to human constraints increases from $-23$ to $7$. The naturalness improvements come from the common constraint synthesizing method, which leads to the constraints that are more natural than human constraints. 

The examples in \Cref{tb:common_examples} show that the constraints from \underline{Single+None} tend to be too specific and the constraints from \underline{Common+None} are often more natural. However, some common constraint like \textit{``Include detailed accounts of the middle portion of events''} is trivially satisfiable and needs revision.
%1. Compared to \textbf{Single+None} baseline, \textbf{Common+Eval} is more challenging and natural.

%\todo{talk about coherence somewhere}

\headingexp{Evaluation-based revision increases the difficulty without degrading the naturalness significantly.} Evaluation-based revision removes those trivially-satisfiable constraints, which are often satisfied in the article summary. Thus, the \underline{Eval} drops the score of both Org and Core. Compared to \underline{Eval}, \underline{Direct}, the baseline that revises all constraints, further increases the difficulty while sacrificing the naturalness. 

%The trivially-satisfiable constraints are often satisf
%Lots of core satisfaction is done trivially.
%Direct revision
%also drops the satisfaction rate after summary 

\headingexp{Common constraints mitigate superficial constraint satisfaction} while being similarly challenging after being combined with evaluation-based revision. 
\Cref{tb:baselines_sat} indicates that the common constraints alone make the constraints easier (e.g., $89.7$ from \underline{Single+None} vs $95.1$ from \underline{Common+None} for \textit{GPT-5 Mini}). Nevertheless, after revision, original satisfaction rates of \underline{Single+Eval} and \underline{Common+Eval} are similar and \textit{GPT-5 Mini} achieves a much higher core satisfaction rate, $46.5$, in \underline{Common+Eval} compared to $35.5$ in \underline{Single+Eval}. This suggests that the more general constraints allow the strong LLMs to satisfy them in their core narratives. 

The \underline{Common} also reliably reduces similarities between the reference article(s) and the constraints in \Cref{tb:ref_sim}. It also reduces the similarities between the generated article and the constraints in \Cref{tb:baselines_sim}, which further supports our conclusion. Finally, simply adding general instruction into the synthesis prompt actually increases the performance gap of \textit{GPT-5 Mini} between the original satisfaction rate and core rate, suggesting encouragement of the copying behavior.

%simply asking LLM to synthesize general constraints cannot maintain their difficulty.

%constraints
%Original response satisfaction rate is slightly lower but the problem is fixed after the evaluation-based revision.

%\Cref{tb:human_exp} 

%remember to also present the result that just ask LLM to output general instruction (make the instruction much easier)
%\subsection{Evaluation-based Revision Validation}

\subsection{Copy-and-Paste Behavior Analyses}

From the experiment results, we can derive the following conclusions:

\headingexp{Evaluation after summarization is a more reliable metric than the similarity measurement.} \Cref{tb:baselines_sat} consistently shows that the core satisfactions rate of \textit{GPT-5 Mini} is much better than \textit{Llama-3 8B} in all settings. Although \Cref{tb:baselines_sim} suggests that the similarity between \textit{Llama-3 8B}'s articles and constraints are on average higher than similarity from \textit{GPT-5 Mini}, which means Llama copies constraints more, the difference is small and in a few settings, \textit{GPT-5 Mini} has a higher similarity. This highlights that the similarity metrics are easily affected by many other factors such as the text style and cannot become a reliable reward function. \Cref{tb:main_sim} in the appendix can also support the finding.%\todo{check position}

\headingexp{Editing an article is more likely to induce the superficial constraint satisfaction} from a strong LLM than generating the article from scratch. Editing an article makes the task more difficult by adding a constraint that the output article should be similar to the base article. Although \textit{GPT-5 Mini} can maintain similar original constraint satisfaction rate, its core rate significantly decreases in \Cref{tb:baselines_sat}, which indicates more constraints are satisfied superficially.  
%\Cref{tb:baselines_sim}
%edit is more difficult to an weaker LLM like \textit{Llama-3 8B}

% \begin{table*}[t]
%     \centering
%     \scalebox{0.85}{
%     \begin{tabular}{c|cccc|cccc|cccc}
%         \toprule
%          & \multicolumn{4}{c|}{News} & \multicolumn{4}{c|}{Story} & \multicolumn{4}{c}{Blog} \\
%          & \multicolumn{2}{c}{Single} & \multicolumn{2}{c|}{Common} & \multicolumn{2}{c}{Single} & \multicolumn{2}{c|}{Common} & \multicolumn{2}{c}{Single} & \multicolumn{2}{c}{Common} \\
%          & Org & Core  & Org & Core  &  Org & Core & Org & Core  & Org & Core  &  Org & Core   \\
%          \midrule
%          Claude Haiku 4.5 &  &  &  &  & & \\
%          GPT-5 Mini &  &  &  &  & & \\
%          GPT-5 Nano & 68.1  & 14.3  & 70.6  & 19.3 &  &\\
%          LFM2 24B A2B&  &  &  &  &  & \\
%          Gemma-3n E4B &  &  &  &  &  &\\
%          Qwen2.5 7B Turbo&  &  &  &  &  &\\
%          Laama-3 8B Lite&  &  &  &  &  &\\
%          \bottomrule
%     \end{tabular}
%     }
%     \caption{The average satisfaction rate from Haiku 4.5 + GPT-5 Nano. The numbers are percentage.}
%     \label{tb:main_haiku_gpt}
% \end{table*}
\begin{table*}[t]
    \centering
    \scalebox{0.85}{
    \begin{tabular}{c|cccc|cccc|cccc}
        \toprule
         & \multicolumn{4}{c|}{News} & \multicolumn{4}{c|}{Story} & \multicolumn{4}{c}{Blog} \\
         & \multicolumn{2}{c}{Single} & \multicolumn{2}{c|}{Common} & \multicolumn{2}{c}{Single} & \multicolumn{2}{c|}{Common} & \multicolumn{2}{c}{Single} & \multicolumn{2}{c}{Common} \\
         & Org & Core  & Org & Core  &  Org & Core & Org & Core  & Org & Core  &  Org & Core   \\
         \midrule
         Claude Sonnet 4.5 & 89.6 & 15.9 & 85.2 & 22.7 & 89.6 & 25.9 & 91.7 & 44.4 & 82.9 & 39.2 & 87.2 & 50.9 \\
         Claude Haiku 4.5 & 74.7 & 16.0 & 71.8 & 19.1 & 83.8 & 26.9 & 85.9 & 41.1 & 74.8 & 36.8 & 78.0 & 45.9 \\
         GPT-5 Mini & 79.0 & 30.6 & 78.5 & 36.1 & 94.7 & 57.4 & 95.1 & 69.9 & 85.9 & 45.6 & 84.3 & 55.1 \\
         GPT-5 Nano & 68.5 & 15.7 & 69.5 & 17.8 & 81.4 & 31.4 & 86.2 & 47.3 & 74.6 & 36.0 & 75.5 & 49.1 \\
         LFM2 24B A2B& 58.1 & 12.6 & 61.5 & 14.4 & 70.7 & 24.9 & 79.0 & 42.5 & 56.8 & 30.8 & 69.9 & 41.5 \\
         Gemma-3n E4B & 59.3 & 12.1 & 56.5 & 15.5 & 76.0 & 26.4 & 77.8 & 40.3 & 57.5 & 28.1 & 56.2 & 35.4 \\
         Qwen2.5 7B Turbo& 54.6 & 13.1 & 51.3 & 15.4 & 66.0 & 19.4 & 58.5 & 25.4 & 55.0 & 30.4 & 58.1 & 38.1 \\
         Llama-3.3 70B Lite & 60.5 & 13.7 & 51.4 & 13.7 & 69.5 & 25.1 & 64.5 & 30.0 & 51.9 & 30.1 & 49.0 & 32.0 \\
         Llama-3 8B Lite& 34.2 & 8.3 & 33.9 & 6.8 & 45.2 & 19.1 & 50.5 & 26.2 & 34.9 & 22.4 & 42.4 & 27.4 \\
         \bottomrule
    \end{tabular}
    }
    \caption{The average satisfaction rate from Haiku 4.5 + GPT-5 Nano in \methodname. The numbers are percentages. Maximum standard error across all cells is 3.0.}
    \label{tb:main_haiku_gpt}
\end{table*}
\begin{table*}[t]
    \centering
    \scriptsize
    \setlength{\tabcolsep}{3pt}
    \resizebox{\textwidth}{!}{
    \begin{tabular}{c|cccccccc|cccccccc|cccccccc}
        \toprule
         & \multicolumn{8}{c|}{News} & \multicolumn{8}{c|}{Story} & \multicolumn{8}{c}{Blog} \\
         & \multicolumn{4}{c}{Single} & \multicolumn{4}{c|}{Common} & \multicolumn{4}{c}{Single} & \multicolumn{4}{c|}{Common} & \multicolumn{4}{c}{Single} & \multicolumn{4}{c}{Common} \\
         & \multicolumn{2}{c}{Org} & \multicolumn{2}{c}{Core} & \multicolumn{2}{c}{Org} & \multicolumn{2}{c|}{Core} & \multicolumn{2}{c}{Org} & \multicolumn{2}{c}{Core} & \multicolumn{2}{c}{Org} & \multicolumn{2}{c|}{Core} & \multicolumn{2}{c}{Org} & \multicolumn{2}{c}{Core} & \multicolumn{2}{c}{Org} & \multicolumn{2}{c}{Core} \\
         & Coh & Like & Coh & Like & Coh & Like & Coh & Like & Coh & Like & Coh & Like & Coh & Like & Coh & Like & Coh & Like & Coh & Like & Coh & Like & Coh & Like \\
         \midrule
         Claude Sonnet 4.5 & 3.46 & 3.44 & 2.92 & 2.56 & 3.56 & 3.50 & 2.62 & 2.42 & 3.50 & 3.98 & 2.77 & 2.67 & 3.58 & 4.00 & 2.60 & 2.56 & 3.12 & 3.12 & 2.76 & 2.58 & 3.48 & 3.54 & 3.02 & 2.98 \\
         Claude Haiku 4.5 & 3.50 & 3.50 & 3.12 & 2.52 & 3.66 & 3.36 & 2.82 & 2.46 & 3.58 & 3.92 & 2.56 & 2.58 & 3.70 & 3.82 & 2.62 & 2.56 & 3.20 & 3.14 & 2.93 & 2.70 & 3.50 & 3.68 & 2.92 & 2.86 \\
         GPT-5 Mini & 3.32 & 3.00 & 3.02 & 2.50 & 3.28 & 3.24 & 2.88 & 2.44 & 3.24 & 3.70 & 2.82 & 3.36 & 3.32 & 3.75 & 2.72 & 3.30 & 3.08 & 2.78 & 2.66 & 2.32 & 3.32 & 3.64 & 2.72 & 2.68 \\
         GPT-5 Nano & 2.92 & 3.02 & 2.74 & 2.36 & 3.38 & 3.18 & 2.66 & 2.24 & 2.56 & 3.18 & 2.38 & 2.54 & 2.78 & 3.34 & 2.50 & 2.58 & 2.76 & 2.54 & 2.62 & 2.34 & 3.28 & 3.36 & 2.72 & 2.62 \\
         LFM2 24B A2B & 1.90 & 2.00 & 2.66 & 2.34 & 1.98 & 2.24 & 2.42 & 2.26 & 1.76 & 2.18 & 2.46 & 2.54 & 1.90 & 2.28 & 2.26 & 2.48 & 1.90 & 1.94 & 2.56 & 2.44 & 2.16 & 2.44 & 2.60 & 2.64 \\
         Gemma-3n E4B & 2.92 & 3.06 & 2.84 & 2.24 & 2.76 & 2.86 & 2.66 & 2.22 & 2.72 & 3.22 & 2.56 & 2.52 & 3.12 & 3.64 & 2.48 & 2.44 & 2.82 & 2.74 & 2.70 & 2.46 & 2.68 & 3.04 & 2.50 & 2.64 \\
         Qwen2.5 7B Turbo & 2.50 & 2.58 & 2.88 & 2.32 & 2.58 & 2.66 & 2.50 & 2.04 & 2.34 & 2.44 & 2.60 & 2.30 & 2.52 & 2.66 & 2.44 & 2.22 & 2.54 & 2.34 & 2.66 & 2.28 & 2.64 & 2.72 & 2.60 & 2.48 \\
         Llama-3.3 70B Lite & 2.42 & 2.38 & 2.54 & 2.18 & 2.48 & 2.56 & 2.54 & 2.14 & 2.52 & 2.74 & 2.40 & 2.38 & 2.54 & 2.60 & 2.48 & 2.40 & 2.54 & 2.32 & 2.56 & 2.36 & 2.60 & 2.68 & 2.66 & 2.59 \\
         Llama-3 8B Lite & 2.31 & 2.38 & 2.52 & 2.10 & 2.28 & 2.18 & 2.12 & 1.82 & 2.16 & 2.20 & 2.08 & 2.08 & 2.36 & 2.38 & 2.34 & 2.30 & 2.46 & 2.30 & 2.68 & 2.38 & 2.54 & 2.48 & 2.64 & 2.50 \\
         \bottomrule
    \end{tabular}
    }
    \caption{The average quality scores from Haiku 4.5 + GPT-5 Nano. The Coh is coherence and Like is the likability of the response. The maximum approximate SE across all cells is $0.13$.}
    \label{tb:quality}
\end{table*}

\section{\methodname Benchmark}
In this section, we use our proposed constraint synthesis and evaluation method to build a new benchmark, and compare the performances of LLMs in the benchmark. 

%news, blog, story
%story 

\subsection{Benchmark Construction}
To observe whether our finding holds in the domains other than news, we apply our methods to blogs and stories. The stories are randomly sampled from WritingPrompt dataset~\citep{fan2018hierarchical}. The blogs come from the falcon-refined-web subset of Dolma v1.7 \cite{dolma}, which is a filtered and deduplicated extract of CommonCrawl web pages. We then apply URL-based heuristics to classify a document, marking a document as a blog if its URL resolves to a known blogging platform (e.g., WordPress, Medium, Blogspot), contains a blog-style subdomain, or includes a /blog path segment.

%\todo{Jeet: where do the blog, and story come from? how do the retrieval in the common constraint work?}

%We have gathered the blog dataset from Dolma v1.7 \cite{dolma},  a pretraining corpus released by the Allen Institute for AI. We used a subset named falcon-refined-web, which is a filtered and deduplicated extract of CommonCrawl web pages. We then applied URL-based heuristics to classify a document, marking a document as a blog if its URL resolves to a known blogging platform (e.g., WordPress, Medium, Blogspot), contains a blog-style subdomain, or includes a /blog path segment. Finally, we restricted the candidate pool by length, keeping only blogs between 800 and 2,000 words to exclude fragments and unusually long documents.

Besides our proposed \underline{Common+Eval}, we also evaluate LLMs using \underline{Single+Eval}, which produces more specific constraints, in \methodname to analyze the LLM performance differences to different constraint specificity.
%\underline{Single+Eval} would generate more specific constraints, so in \methodname benchmark, we test both to observe LLMs performance difference to different constraint specificity.

To reduce the evaluation cost of our benchmark, we ask the models to only output yes or no for constraint satisfaction without a reason and retry if their output does not follow our format. We also compare the evaluation results of \textit{Claude Sonnet 4.5}, \textit{Claude Haiku 4.5}, and \textit{GPT-5 Nano} with low reasoning effort in \Cref{app:judge_comp}. We found that the scores from \textit{Sonnet 4.5} and \textit{Haiku 4.5} are very similar. This shows that a smaller LLM can judge the constraint satisfaction reliably.

\textit{GPT-5 Nano} is generally stricter in terms of constraint satisfaction but gives a higher coherence score. \textit{GPT-5 Nano} prefers the articles of the OpenAI models just like \textit{Claude Haiku 4.5} is biased toward Anthropic models~\citep{panickssery2024llm}. We average the scores from \textit{GPT-5 Nano} and \textit{Claude Haiku 4.5} to mitigate the self preference.

%The results confirm that we can use Haiku and Nano to evaluate the results inexpensively.

%to reduce the self bias 
%\textit{GPT-5 Nano} (reasoning effort low)
%\textit{Claude 4.5 Haiku}
%Yes/No only

%\todo{We compare the different judges in XXX}
%Using \underline{Single+Eval} and \underline{Common+Eval}

\subsection{LLM Comparisons}
\label{sec:llm_comp}

We summarize our findings below.

\headingexp{Our diverse metrics could discover different weaknesses in LLMs.} For instance, in \Cref{tb:main_haiku_gpt}, the improvements of \textit{Gemma 3n E4B} over \textit{Llama3.3 70B} in the story and blog domains mostly come from superficial instruction following due to their similar core satisfaction rates. Another example is that \textit{LFM2-24B} and \textit{Gemma 3n E4B} perform similarly for specific constraints from \underline{Single} while \textit{Gemma 3n E4B} performs significantly worse for general constraints from \underline{Common}.

%Core satisfaction rate and original satisfaction rate are also different. 
%The satisfaction rates of \underline{Single} and \underline{Common} are different. For instance, 

\headingexp{LLM tends to satisfy the constraints in news more superficially than in blogs.} The original satisfaction ratios of news are similar to those of blogs in \Cref{tb:main_haiku_gpt}, while having much smaller core satisfaction rates. Among the satisfied constraints, only $28.4$\% of them in news and $63.1$\% of the them in blog are still satisfied after the summarization on average.
Our preliminary investigation shows that it is because news has more facts than blog and the constraints related to the facts are more likely to be satisfied in the details. %\todo{Balpreet: check this}

Interestingly, \Cref{tb:main_sim} in the appendix demonstrates that the responses from news are less similar to the constraints than those from blogs. We hypothesize that this is because news mentions more entities and entity differences reduce the similarity measurement of the sentence BERT model, which supports our motivation of proposing the new evaluation after summarization metrics to analyze the superficial instruction following behavior. %\todo{Balpreet: check this. consider to move this to appendix}

%core is more reliable than similarity
%different names in news might make the similarity very different.
%news constraints are not likely to be copied than blog and story

\headingexp{Some LLMs satisfy more constraints while sacrificing some coherence.} For example, among the open source models, \textit{LFM2 24B} achieves the best original constraint satisfaction rates overall in \Cref{tb:main_haiku_gpt} while having the worst original coherence in \Cref{tb:quality}.
%\todo{check the coherence numbers in different domains again}

\headingexp{Superficial instruction following does not explain all the performance improvement of stronger models.} Stronger models usually achieve high satisfaction rates both in the original article and in its summary in \Cref{tb:main_haiku_gpt}.
%Stronger models follow the instructions better 
%also usually satisfy more constraints after summary.}
%\Cref{tb:main_haiku_gpt} shows that the original satisfaction rate is highly correlated with 
%not all the improvement comes from copy and paste.

\headingexp{As the model size increases, the improvement of some models mostly comes from superficial instruction following.} In \Cref{tb:main_haiku_gpt}, Claude and Llama model families improve the original satisfaction rate much more than the core satisfaction rate while the GPT model series improve core satisfaction rate more. This shows that the copy-and-paste behaviors could be affected by model training recipes and sizes.

%\headingexp{As the model size increases, nearly none of the improvements come from copy and paste.} For example, in \Cref{tb:main_haiku_gpt}, \underline{GPT-5 mini} improves 16.2\% on core satisfaction rate over \underline{GPT-5 Nano}, while only improves 10.3\% on original satisfaction rate. The results also show that the smaller models have more serious copy-and-paste issue.\todo{need more experiments}

%larger models remember more ways to satisfy the constraints

\section{Related Work}

In addition to back-translation, the constraints could also come from decomposing the user instructions~\citep{ferraz2024llm, qin2024infobench, lior2025wildifeval}. However, not many users are willing to type lots of constraints manually in their instructions, so it is difficult to acquire the complex and high-quality user instructions in an arbitrary domain. \citet{guo2025recast, hui2025decif} propose to add existing constraints to instruction to increase the complexity of the instruction, but the consistency check for the new constraints would be increasingly difficult as the instruction becomes more complicated.

%. Although the methods try to check the consistency of the resulting  it is 

%the 
%user usually won't type lots of constraints manually
%approach cannot 
%hard to collect lots of constraints in an arbitrary domain
%also be added to instruction 

%add existing constraints and check consistency
%\cite{guo2025recast, hui2025decif}
%hard to make sure the set of constraints make sense and not contradicting each other

%identify constraints from the instructions from real users or human annotators 

%backtranslation
%~\citep{li2024self, pham-etal-2024-suri, atmakuru2024cs4, qi2025constraint}
%too specific and easy

Our constraints come from multiple documents, which are related to the long-context benchmarks that ask the multi-document questions~\citep{wang2024leave,zhang2024bench,liu2025mdcure}. One major difference is that the target LLMs do not need to access the documents to follow our instruction. Our common constraint synthesis methodology is most similar to \citet{chang2024fine}, which synthesizes the common summary using two documents. However, they only synthesize very short and simple common summary to improve the interactive story generation.

Finally, several instruction synthesis methods have a step that makes instruction more difficult by revision or iteratively adding more constraints ~\cite{wen2024benchmarking,xu2024wizardlm,zhang2025iopo,an2025ultraif,dong2025revisiting,huang2026steerable}, but the steps are more similar to our direct revision baseline, which does not consider the naturalness of the constraints.

%\cite{an2025ultraif}
%iteratively synthesize more constraints
%but not consider the naturalness of the constraints

%Synthesizing instructions from multiple documents is 

%copy and paste behavior
%\cite{cahoon2026clinical}
%different from memorization

%preventing memorization is difficult
%\cite{ippolito2023preventing}

\section{Conclusion}
We propose \methodname, which is composed of three novel interlocking techniques: common constraint synthesis, evaluation-based constraint revision, and satisfaction evaluation after summary. Synthesizing common constraints increases the naturalness of the constraints and reduces copy-and-paste behavior; evaluation-based revision increases the difficulty; evaluation after summary is a reliable tool for analyzing LLMs' copy-and-paste behavior in the instruction following task.

Our extensive experiments demonstrate that copy-and-paste behaviors are influenced by constraint synthesis methods, the text domains, model sizes, and which LLM we used to generate the article. For example, the strong proprietary LLM family such as \textit{GPT-5} would choose to take more copy-and-paste shortcuts when the model is smaller or it is working on a more difficult editing task. In contrast, Claude 4.5 and LLama-3 copy more as model size increases.

%three novel techniques. 

%common constraints improve the instruction following after the summarization
%similarity

%backup strategy
%when the task becomes difficult to the LLM, the LLMs tend to leverage this shortcuts more

%editing
%copy and paste problem is especially serious in smaller models

%evaluation-based revision balances the naturalness and difficulty

%As the model size increases, nearly none of the improvements come from copy and paste.

\section*{Limitations}
Due to our budget limitation, we did not test expensive models with large reasoning budgets such as \textit{GPT-5.5 Pro} or \textit{Claude Opus 4.8}. We also have not used our metrics as the reward signal to see if our method could directly reduce the superficial instruction following. After evaluation-based revision, there might be a small number of constraints that contradict with each other. Finally, we conduct experiments in English corpus and do not know if our conclusions still hold in other languages.

%and there might not be a solution exists

%did not consider the diversity of the generation

\section*{Acknowledgments}
This work was supported in part by the Center for Intelligent Information Retrieval, in part by the National Science Foundation grant \#2106391, and in part by Cisco. Any opinions, findings and conclusions or recommendations expressed in this material are those of the authors and do not necessarily reflect those of the sponsor.

\bibliography{custom}

@inproceedings{fan2018hierarchical,
  title={Hierarchical neural story generation},
  author={Fan, Angela and Lewis, Mike and Dauphin, Yann},
  booktitle={Proceedings of the 56th Annual Meeting of the Association for Computational Linguistics (Volume 1: Long Papers)},
  pages={889--898},
  year={2018}
}

@article{dolma,
  title = {{Dolma: an Open Corpus of Three Trillion Tokens for Language Model Pretraining Research}},
  author={
    Luca Soldaini and Rodney Kinney and Akshita Bhagia and Dustin Schwenk and David Atkinson and
    Russell Authur and Ben Bogin and Khyathi Chandu and Jennifer Dumas and Yanai Elazar and
    Valentin Hofmann and Ananya Harsh Jha and Sachin Kumar and Li Lucy and Xinxi Lyu and
    Nathan Lambert and Ian Magnusson and Jacob Morrison and Niklas Muennighoff and Aakanksha Naik and
    Crystal Nam and Matthew E. Peters and Abhilasha Ravichander and Kyle Richardson and Zejiang Shen and
    Emma Strubell and Nishant Subramani and Oyvind Tafjord and Pete Walsh and Luke Zettlemoyer and
    Noah A. Smith and Hannaneh Hajishirzi and Iz Beltagy and Dirk Groeneveld and Jesse Dodge and Kyle Lo
  },
  year = {2024},
  journal={arXiv preprint},
}

@inproceedings{reimers2019sentence,
  title={Sentence-bert: Sentence embeddings using siamese bert-networks},
  author={Reimers, Nils and Gurevych, Iryna},
  booktitle={Proceedings of the 2019 conference on empirical methods in natural language processing and the 9th international joint conference on natural language processing (EMNLP-IJCNLP)},
  pages={3982--3992},
  year={2019}
}

@inproceedings{ye2026muldimif,
  title={MulDimIF: A Multi-Dimensional Constraint Framework for Evaluating and Improving Instruction Following in Large Language Models},
  author={Ye, Junjie and Huang, Caishuang and Chen, Zhuohan and Fu, Wenjie and Yang, Chenyuan and Yang, Leyi and Wu, Yilong and Wang, Peng and Zhou, Meng and Yang, Xiaolong and others},
  booktitle={Findings of the Association for Computational Linguistics: ACL 2026},
  pages={2078--2104},
  year={2026}
}

@article{panickssery2024llm,
  title={Llm evaluators recognize and favor their own generations},
  author={Panickssery, Arjun and Bowman, Samuel R and Feng, Shi},
  journal={Advances in Neural Information Processing Systems},
  volume={37},
  pages={68772--68802},
  year={2024}
}

@article{dong2025revisiting,
  title={Revisiting the Reliability of Language Models in Instruction-Following},
  author={Dong, Jianshuo and Zhang, Yutong and Liu, Yan and Zhong, Zhenyu and Wei, Tao and Zhang, Chao and Qiu, Han},
  journal={arXiv preprint arXiv:2512.14754},
  year={2025}
}

@inproceedings{zellers2019hellaswag,
    title={HellaSwag: Can a Machine Really Finish Your Sentence?},
    author={Zellers, Rowan and Holtzman, Ari and Bisk, Yonatan and Farhadi, Ali and Choi, Yejin},
    booktitle ={Proceedings of the 57th Annual Meeting of the Association for Computational Linguistics},
    year={2019}
}

@inproceedings{zhang2024bench,
  title={Infinity Bench: Extending long context evaluation beyond 100K tokens},
  author={Zhang, Xinrong and Chen, Yingfa and Hu, Shengding and Xu, Zihang and Chen, Junhao and Hao, Moo and Han, Xu and Thai, Zhen and Wang, Shuo and Liu, Zhiyuan and others},
  booktitle={Proceedings of the 62nd Annual Meeting of the Association for Computational Linguistics (Volume 1: Long Papers)},
  pages={15262--15277},
  year={2024}
}

@inproceedings{wang2024leave,
  title={Leave no document behind: Benchmarking long-context llms with extended multi-doc qa},
  author={Wang, Minzheng and Chen, Longze and Cheng, Fu and Liao, Shengyi and Zhang, Xinghua and Wu, Bingli and Yu, Haiyang and Xu, Nan and Zhang, Lei and Luo, Run and others},
  booktitle={Proceedings of the 2024 Conference on Empirical Methods in Natural Language Processing},
  pages={5627--5646},
  year={2024}
}

@inproceedings{chen2024dog,
  title={Dog-instruct: Towards premium instruction-tuning data via text-grounded instruction wrapping},
  author={Chen, Yongrui and Jiang, Haiyun and Huang, Xinting and Shi, Shuming and Qi, Guilin},
  booktitle={Proceedings of the 2024 Conference of the North American Chapter of the Association for Computational Linguistics: Human Language Technologies (Volume 1: Long Papers)},
  pages={4125--4135},
  year={2024}
}

@inproceedings{nguyen2024better,
  title={Better alignment with instruction back-and-forth translation},
  author={Nguyen, Thao and Li, Jeffrey and Oh, Sewoong and Schmidt, Ludwig and Weston, Jason E and Zettlemoyer, Luke and Li, Xian},
  booktitle={Findings of the Association for Computational Linguistics: EMNLP 2024},
  pages={13289--13308},
  year={2024}
}

@inproceedings{liu2025mdcure,
  title={Mdcure: A scalable pipeline for multi-document instruction-following},
  author={Liu, Gabrielle Kaili-May and Shi, Bowen and Caciularu, Avi and Szpektor, Idan and Cohan, Arman},
  booktitle={Proceedings of the 63rd Annual Meeting of the Association for Computational Linguistics (Volume 1: Long Papers)},
  pages={29258--29296},
  year={2025}
}

@inproceedings{chang2024fine,
  title={Fine-to-coarse entailment hierarchy construction for coarse-to-fine story generation},
  author={Chang, Haw-Shiuan and Peng, Nanyun and Bansal, Mohit and Chung, Tagyoung},
  booktitle = {Bridging Human-Computer Interaction and Natural Language Processing (HCI+NLP)},
  year={2024}
}

@inproceedings{ippolito2023preventing,
  title={Preventing Generation of Verbatim Memorization in Language Models Gives a False Sense of Privacy},
  author={Ippolito, Daphne and Tramer, Florian and Nasr, Milad and Zhang, Chiyuan and Jagielski, Matthew and Lee, Katherine and Choo, Christopher Choquette and Carlini, Nicholas},
  booktitle={Proceedings of the 16th International Natural Language Generation Conference},
  pages={28--53},
  year={2023}
}

@inproceedings{qin2024infobench,
  title={Infobench: Evaluating instruction following ability in large language models},
  author={Qin, Yiwei and Song, Kaiqiang and Hu, Yebowen and Yao, Wenlin and Cho, Sangwoo and Wang, Xiaoyang and Wu, Xuansheng and Liu, Fei and Liu, Pengfei and Yu, Dong},
  booktitle={Findings of the Association for Computational Linguistics: ACL 2024},
  pages={13025--13048},
  year={2024}
}

@article{lior2025wildifeval,
  title={Wildifeval: Instruction following in the wild},
  author={Lior, Gili and Yehudai, Asaf and Gera, Ariel and Ein-Dor, Liat},
  journal={arXiv preprint arXiv:2503.06573},
  year={2025}
}

@inproceedings{ferraz2024llm,
  title={LLM self-correction with DeCRIM: Decompose, critique, and refine for enhanced following of instructions with multiple constraints},
  author={Ferraz, Thomas Palmeira and Mehta, Kartik and Lin, Yu-Hsiang and Chang, Haw-Shiuan and Oraby, Shereen and Liu, Sijia and Subramanian, Vivek and Chung, Tagyoung and Bansal, Mohit and Peng, Nanyun},
  booktitle={Findings of the Association for Computational Linguistics: EMNLP 2024},
  pages={7773--7812},
  year={2024}
}

@inproceedings{zhang2025iopo,
  title={Iopo: Empowering llms with complex instruction following via input-output preference optimization},
  author={Zhang, Xinghua and Yu, Haiyang and Fu, Cheng and Huang, Fei and Li, Yongbin},
  booktitle={Proceedings of the 63rd Annual Meeting of the Association for Computational Linguistics (Volume 1: Long Papers)},
  pages={22185--22200},
  year={2025}
}

@article{hui2025decif,
  title={Decif: Improving instruction-following through meta-decomposition},
  author={Hui, Tingfeng and Zhu, Pengyu and Ping, Bowen and Tang, Ling and Dong, Guanting and Zhang, Yaqi and Su, Sen},
  journal={arXiv preprint arXiv:2505.13990},
  year={2025}
}

@inproceedings{an2025ultraif,
  title={UltraIF: Advancing Instruction Following from the Wild},
  author={An, Kaikai and Sheng, Li and Cui, Ganqu and Si, Shuzheng and Ding, Ning and Cheng, Yu and Chang, Baobao},
  booktitle={Proceedings of the 2025 Conference on Empirical Methods in Natural Language Processing},
  pages={18722--18737},
  year={2025}
}

@inproceedings{wang2023self,
  title={Self-instruct: Aligning language models with self-generated instructions},
  author={Wang, Yizhong and Kordi, Yeganeh and Mishra, Swaroop and Liu, Alisa and Smith, Noah A and Khashabi, Daniel and Hajishirzi, Hannaneh},
  booktitle={Proceedings of the 61st annual meeting of the association for computational linguistics (volume 1: long papers)},
  pages={13484--13508},
  year={2023}
}

@inproceedings{li2024self,
  title={Self-alignment with instruction backtranslation},
  author={Li, Xian and Yu, Ping and Zhou, Chunting and Schick, Timo and Levy, Omer and Zettlemoyer, Luke and Weston, Jason E and Lewis, Mike},
  booktitle={International Conference on Learning Representations},
  volume={2024},
  pages={3552--3577},
  year={2024}
}

@article{atmakuru2024cs4,
  title={Cs4: Measuring the creativity of large language models automatically by controlling the number of story-writing constraints},
  author={Atmakuru, Anirudh and Nainani, Jatin and Bheemreddy, Rohith Siddhartha Reddy and Lakkaraju, Anirudh and Yao, Zonghai and Zamani, Hamed and Chang, Haw-Shiuan},
  journal={arXiv preprint arXiv:2410.04197},
  year={2024}
}

@inproceedings{lu2025benchmarking,
  title={Benchmarking language model creativity: A case study on code generation},
  author={Lu, Yining and Wang, Dixuan and Li, Tianjian and Jiang, Dongwei and Khudanpur, Sanjeev and Jiang, Meng and Khashabi, Daniel},
  booktitle={Proceedings of the 2025 Conference of the Nations of the Americas Chapter of the Association for Computational Linguistics: Human Language Technologies (Volume 1: Long Papers)},
  pages={2776--2794},
  year={2025}
}

@inproceedings{zhang2025cfbench,
  title={Cfbench: A comprehensive constraints-following benchmark for llms},
  author={Zhang, Tao and Zhu, Chenglin and Shen, Yanjun and Luo, Wenjing and Zhang, Yan and Liang, Hao and Yang, Fan and Lin, Mingan and Qiao, Yujing and Chen, Weipeng and others},
  booktitle={Proceedings of the 63rd Annual Meeting of the Association for Computational Linguistics (Volume 1: Long Papers)},
  pages={32926--32944},
  year={2025}
}

@inproceedings{jaroslawicz2025many,
  title={How Many Instructions Can LLMs Follow at Once?},
  author={Jaroslawicz, Daniel and Whiting, Brendan and Shah, Parth and Maamari, Karime},
  booktitle={NeurIPS 2025 Workshop on Evaluating the Evolving LLM Lifecycle: Benchmarks, Emergent Abilities, and Scaling},
  year={2025}
}

@inproceedings{pham-etal-2024-suri,
    title = "{S}uri: Multi-constraint Instruction Following in Long-form Text Generation",
    author = "Pham, Chau Minh  and
      Sun, Simeng  and
      Iyyer, Mohit",
    editor = "Al-Onaizan, Yaser  and
      Bansal, Mohit  and
      Chen, Yun-Nung",
    booktitle = "Findings of the Association for Computational Linguistics: EMNLP 2024",
    month = nov,
    year = "2024",
    address = "Miami, Florida, USA",
    publisher = "Association for Computational Linguistics",
    url = "https://aclanthology.org/2024.findings-emnlp.94/",
    doi = "10.18653/v1/2024.findings-emnlp.94",
    pages = "1722--1753",
}

@inproceedings{qi2025constraint,
  title={Constraint back-translation improves complex instruction following of large language models},
  author={Qi, Yunjia and Peng, Hao and Wang, Xiaozhi and Xu, Bin and Hou, Lei and Li, Juanzi},
  booktitle={Proceedings of the 34th ACM International Conference on Information and Knowledge Management},
  pages={2388--2398},
  year={2025}
}

@article{sun2024conifer,
  title={Conifer: Improving complex constrained instruction-following ability of large language models},
  author={Sun, Haoran and Liu, Lixin and Li, Junjie and Wang, Fengyu and Dong, Baohua and Lin, Ran and Huang, Ruohui},
  journal={arXiv preprint arXiv:2404.02823},
  year={2024}
}

@article{wen2024benchmarking,
  title={Benchmarking complex instruction-following with multiple constraints composition},
  author={Wen, Bosi and Ke, Pei and Gu, Xiaotao and Wu, Lindong and Huang, Hao and Zhou, Jinfeng and Li, Wenchuang and Hu, Binxin and Gao, Wendy and Xu, Jiaxin and others},
  journal={Advances in Neural Information Processing Systems},
  volume={37},
  pages={137610--137645},
  year={2024}
}

@article{huang2026steerable,
  title={Steerable Instruction Following Coding Data Synthesis with Actor-Parametric Schema Co-Evolution},
  author={Huang, Tinglin and Chen, Bo and Zhang, Xiao and Shen, Kai and Ying, Rex},
  journal={arXiv preprint arXiv:2604.16322},
  year={2026}
}

@inproceedings{xu2024wizardlm,
  title={WizardLM: Empowering large pre-trained language models to follow complex instructions},
  author={Xu, Can and Sun, Qingfeng and Zheng, Kai and Geng, Xiubo and Zhao, Pu and Feng, Jiazhan and Tao, Chongyang and Lin, Qingwei and Jiang, Daxin},
  booktitle={International Conference on Learning Representations},
  volume={2024},
  pages={30745--30766},
  year={2024}
}

@article{guo2025recast,
  title={RECAST: Expanding the Boundaries of LLMs' Complex Instruction Following with Multi-Constraint Data},
  author={Guo, Zhengkang and Liu, Wenhao and Xie, Mingchen and Xu, Jingwen and Huang, Zisu and Tian, Muzhao and Xu, Jianhan and Shen, Yuanzhe and Qian, Qi and Wu, Muling and others},
  journal={arXiv preprint arXiv:2505.19030},
  year={2025}
}

@inproceedings{ariyarathne_nwala_3dlnews,
  author    = {Gangani Ariyarathne and Alexander C. Nwala},
  title     = {3DLNews: A Three-decade Dataset of US Local News Articles},
  booktitle = {Proceedings of the 33rd ACM International Conference on Information and Knowledge Management (CIKM ’24)},
  year      = {2024},
  pages     = {1--5},
  location  = {Boise, ID, USA},
  publisher = {ACM},
  address   = {New York, NY, USA},
  doi       = {10.1145/3627673.3679165},
  url       = {https://doi.org/10.1145/3627673.3679165}
}

\appendix

\clearpage

\begin{table}[t]
    \centering
    \scalebox{0.7}{
    \begin{tabular}{cc|cccccc}
        \toprule
         &  &  \multicolumn{3}{c}{GPT-5 Mini} & \multicolumn{3}{c}{Llama-3 8B} \\
        Domain & Synth & Org & Core & Short & Org & Core & Short \\
         \midrule
        \multirow{2}{*}{News} & Single & 78.8 & 29.3 & 37.1 & 31.2 & 6.5 & 21.2  \\
         & Common & 80.8 & 38.5 & 41.5 & 33.1 & 6.9 & 21.3 \\
        \midrule
        \multirow{2}{*}{Story} & Single & 94.6 & 56.2 & 63.9 & 41.9 & 16.8 & 32.2 \\
         & Common &  95.6 & 71.7 & 76.2 & 49.5 & 27.8 & 42.5 \\
        \midrule
        \multirow{2}{*}{Blog} & Single & 83.2 & 40.1 & 40.8 & 32.2 & 18.4 & 19.9 \\
         & Common & 83.7 & 54.4 & 54.9 & 39.9 & 25.5 & 27.9 \\
         \bottomrule
    \end{tabular}
    }
    \caption{Comparison between original satisfaction rates (Org), core satisfaction rates (Core), and the satisfaction rates of directly generating a short article (Short). The constraint synthesis methods are Single+Eval and Common+Eval. }
    \label{tb:len_base}
\end{table}

\section{More Results}
\label{app:more_results}

In this section, we first investigate how much the low core satisfaction rates could be explained by the article length in \Cref{app:short}, analyze the scores from different LLM judges in \Cref{app:judge_comp}, 
present the satisfaction rates of base article in \Cref{app:base_sat}, and report the similarity in three domains between articles and constraints in \Cref{app:similairty}.

\subsection{Length Factor in Core Satisfaction Metric}
\label{app:short}

To investigate how much the lower core satisfaction rates come from the limited summary length, we conduct one more experiment. Instead of asking LLMs to generate an article with $500$ words and summarizing the article to $125$ words, we directly ask LLMs to generate an article that follows the constraints with $125$ words. We put the satisfaction rates into the \underline{Short} column of \Cref{tb:len_base}. 

For \textit{Llama-3 8B}, the core satisfaction rates are much lower than those of the directly generated article with short length in the news and story domains, which suggests the large portion of core satisfaction rates comes from superficial instruction following rather than the length restriction. For \textit{GPT-5 Mini}, the differences are smaller due to less superficial instruction following behavior. The large gaps of \underline{Single+Eval} compared with \underline{Common+Eval} verify the further mitigation of superficial instruction following using common constraint generation.

Notice that the slightly different numbers of \underline{Org}/\underline{Core} between \Cref{tb:len_base} and \Cref{tb:main_haiku_gpt} come from different random seeds for \textit{GPT-5 Mini}. Besides seed differences, we use the deprecated \textit{Llama-3 8B Lite} models from Together AI\footnote{\url{https://api.together.ai/}} in \Cref{tb:main_haiku_gpt} and \textit{Llama-3 8B} hosted by vLLM in \Cref{tb:len_base}.

\subsection{LLM Judge Comparison}
\label{app:judge_comp}

To know how reliable our LLM judges are, we evaluate the constraint satisfaction using \textit{Claude Sonnet 4.5} in \Cref{tb:main_sonnet}, \textit{Claude Haiku 4.5} in \Cref{tb:main_haiku}, and \textit{GPT-5 Nano} in \Cref{tb:main_gpt}. We can see that the results of \Cref{tb:main_sonnet} and \Cref{tb:main_haiku} are almost the same for both \textbf{Single} and \textbf{Common}. This suggests that \textit{Haiku 4.5} is enough to conduct evaluation and constraints from \textbf{Common} is not more difficult to evaluate. Finally, \textit{GPT-5 Nano} is stricter than \textit{Claude} but the performance rank of different generation LLM is similar.

\subsection{Base Article Satisfaction Rates}
\label{app:base_sat}
Some common constraints are too general and easily satisfied. For example, \Cref{tb:revision_perc} shows that more than $50$\% of common constraints could be satisfied by base articles. This motivates our evaluation-based revision. We can see that constraints from the story domain are especially easy to be satisfied compared with news and blog. The results are consistent with our \Cref{tb:main_haiku_gpt}.

\subsection{Similarities in All Domains}
\label{app:similairty}
The models that copy the constraints more tend to have higher similarity. For example, in \Cref{tb:main_sim}, \textit{Claude Sonnet 4.5} copies more than \textit{Claude Haiku 4.5} and \textit{Llama-3.3 70B} copies more than \textit{Llama-3 8B}, while  \textit{GPT-5 Nano} copies more than \textit{GPT-5 Mini}. This aligns with our conclusion in \Cref{sec:llm_comp}.
%shows the similarity between the constraints and 

%Compare the Claude Sonnet, Haiku, GPT-5 Nano LLM-as-a-Judge results 
%(we are actually making the constraints more difficult rather than making constraints harder to judge)

% \begin{table*}[t]
% \centering
% \small
% \setlength{\tabcolsep}{6pt}
% \renewcommand{\arraystretch}{1.2}
% \begin{tabular}{@{}>{\raggedright\arraybackslash}p{0.17\textwidth}>{\raggedright\arraybackslash}p{0.78\textwidth}@{}}
% \toprule
% \textbf{Source} & \textbf{Text} \\
% \midrule
% Reference Article A & ``\ldots The \textbf{Alaska Association of Student Government} coordinated statewide student walkouts after the governor vetoed the education bill.'' \\
% \addlinespace
% Single Constraint 7 & ``Identify the \textbf{Alaska Association of Student Government} as the organizer of the protest.'' \\
% \addlinespace
% Common Constraint 10 & ``Describe the organizational efforts behind the protests, including the involvement of \textbf{student government or community groups}.'' \\
% \bottomrule
% \end{tabular}
% \caption{The constraint from a single article copies a name from the reference
% article. The common constraint from two articles keeps the same idea without
% the name. Bold marks the copied text and its general version.}
% \label{tab:qual_copy}
% \end{table*}

\begin{table*}[t]
    \centering
    \scalebox{0.85}{
    \begin{tabular}{c|cccc|cccc|cccc}
        \toprule
         & \multicolumn{4}{c|}{News} & \multicolumn{4}{c|}{Story} & \multicolumn{4}{c}{Blog} \\
         & \multicolumn{2}{c}{Single} & \multicolumn{2}{c|}{Common} & \multicolumn{2}{c}{Single} & \multicolumn{2}{c|}{Common} & \multicolumn{2}{c}{Single} & \multicolumn{2}{c}{Common} \\
         & Org & Core  & Org & Core  &  Org & Core & Org & Core  & Org & Core  &  Org & Core   \\
         \midrule
         Claude Sonnet 4.5 & 89.4 & 26.9 & 85.5 & 33.3 & 91.9 & 28.7 & 88.4 & 43.1 & 81.3 & 43.1 & 86.1 & 59.8 \\
         Claude Haiku 4.5 & 76.1 & 22.4 & 72.9 & 28.5 & 81.4 & 26.0 & 80.8 & 41.6 & 75.9 & 40.3 & 76.8 & 51.3 \\
         GPT-5 Mini & 77.2 & 37.7 & 77.5 & 45.7 & 90.8 & 53.2 & 92.5 & 68.0 & 81.3 & 46.3 & 82.7 & 58.9 \\
         GPT-5 Nano & 69.2 & 24.8 & 68.6 & 29.1 & 77.1 & 34.2 & 81.6 & 43.0 & 72.1 & 41.9 & 73.3 & 52.2 \\
         LFM2 24B A2B& 62.9 & 17.5 & 71.6 & 24.6 & 74.9 & 26.2 & 80.4 & 40.8 & 63.2 & 33.6 & 73.1 & 48.7 \\
         Gemma-3n E4B & 62.9 & 18.8 & 58.7 & 21.1 & 71.9 & 31.3 & 71.4 & 40.3 & 57.1 & 29.8 & 58.6 & 39.4 \\
         Qwen2.5 7B Turbo& 53.7 & 19.7 & 51.1 & 20.9 & 59.5 & 20.7 & 54.1 & 26.5 & 49.3 & 31.8 & 53.1 & 39.6 \\
         Llama-3.3 70B Lite & 58.1 & 20.7 & 50.0 & 19.6 & 62.8 & 23.6 & 59.8 & 32.2 & 49.7 & 28.2 & 51.5 & 35.4 \\
         Llama-3 8B Lite& 34.9 & 10.2 & 35.4 & 11.3 & 40.0 & 18.1 & 44.0 & 26.5 & 31.1 & 20.8 & 39.9 & 30.4 \\
         \bottomrule
    \end{tabular}
    }
    \caption{The average satisfaction rate from Claude Sonnet 4.5. All values are reported as percentages. The maximum standard error across all cells is 3.9 percentage points.}
    \label{tb:main_sonnet}
\end{table*}

\begin{table*}[t]
    \centering
    \scalebox{0.85}{
    \begin{tabular}{c|cccc|cccc|cccc}
        \toprule
         & \multicolumn{4}{c|}{News} & \multicolumn{4}{c|}{Story} & \multicolumn{4}{c}{Blog} \\
         & \multicolumn{2}{c}{Single} & \multicolumn{2}{c|}{Common} & \multicolumn{2}{c}{Single} & \multicolumn{2}{c|}{Common} & \multicolumn{2}{c}{Single} & \multicolumn{2}{c}{Common} \\
         & Org & Core  & Org & Core  &  Org & Core & Org & Core  & Org & Core  &  Org & Core   \\
         \midrule
         Claude Sonnet 4.5 & 89.3 & 24.2 & 86.0 & 31.2 & 90.4 & 34.8 & 91.5 & 52.1 & 82.5 & 44.8 & 86.6 & 61.7 \\
         Claude Haiku 4.5 & 76.8 & 24.6 & 74.8 & 28.0 & 85.4 & 37.0 & 88.0 & 49.4 & 74.0 & 45.3 & 78.5 & 54.4 \\
         GPT-5 Mini & 78.8 & 37.3 & 79.1 & 43.3 & 95.9 & 64.9 & 95.7 & 78.3 & 84.3 & 50.3 & 83.5 & 61.1 \\
         GPT-5 Nano & 69.1 & 24.4 & 67.7 & 24.5 & 81.0 & 39.0 & 86.2 & 54.6 & 74.2 & 42.0 & 75.5 & 54.9 \\
         LFM2 24B A2B& 61.4 & 19.7 & 65.4 & 22.0 & 76.4 & 33.5 & 80.9 & 51.0 & 58.8 & 36.3 & 71.4 & 49.6 \\
         Gemma-3n E4B & 61.3 & 18.6 & 58.6 & 22.8 & 78.3 & 36.1 & 80.5 & 48.8 & 58.4 & 34.0 & 59.3 & 41.4 \\
         Qwen2.5 7B Turbo& 53.5 & 19.0 & 50.2 & 20.3 & 65.1 & 28.2 & 59.2 & 31.5 & 51.1 & 33.7 & 56.6 & 41.4 \\
         Llama-3.3 70B Lite & 58.5 & 18.8 & 50.2 & 19.8 & 66.6 & 34.0 & 63.4 & 35.7 & 48.6 & 31.3 & 50.7 & 36.7 \\
         Llama-3 8B Lite& 33.7 & 9.8 & 32.8 & 9.1 & 43.6 & 21.7 & 51.2 & 32.7 & 33.7 & 23.6 & 39.9 & 31.3 \\
         \bottomrule
    \end{tabular}
    }
    \caption{The average satisfaction rate from Haiku 4.5. All values are reported as percentages. The maximum standard error across all cells is 4.1 percentage points.}
    \label{tb:main_haiku}
\end{table*}

\begin{table*}[t]
    \centering
    \scalebox{0.85}{
    \begin{tabular}{c|cccc|cccc|cccc}
        \toprule
         & \multicolumn{4}{c|}{News} & \multicolumn{4}{c|}{Story} & \multicolumn{4}{c}{Blog} \\
         & \multicolumn{2}{c}{Single} & \multicolumn{2}{c|}{Common} & \multicolumn{2}{c}{Single} & \multicolumn{2}{c|}{Common} & \multicolumn{2}{c}{Single} & \multicolumn{2}{c}{Common} \\
         & Org & Core  & Org & Core  &  Org & Core & Org & Core  & Org & Core  &  Org & Core   \\
         \midrule
         Claude Sonnet 4.5 & 89.9 & 7.5 & 84.3 & 14.2 & 88.7 & 16.9 & 91.9 & 36.6 & 83.3 & 33.5 & 87.7 & 40.1 \\
         Claude Haiku 4.5 & 72.6 & 7.3 & 68.7 & 10.2 & 82.3 & 16.8 & 83.8 & 32.9 & 75.8 & 28.4 & 77.4 & 37.4 \\
         GPT-5 Mini & 79.2 & 23.9 & 78.0 & 29.0 & 93.4 & 49.9 & 94.4 & 61.5 & 87.6 & 41.0 & 85.2 & 49.1 \\
         GPT-5 Nano & 68.0 & 7.0 & 71.2 & 10.9 & 81.8 & 23.7 & 86.3 & 40.0 & 74.9 & 30.1 & 75.4 & 43.3 \\
         LFM2 24B A2B& 54.9 & 5.4 & 57.5 & 6.9 & 65.0 & 16.3 & 77.0 & 34.1 & 54.8 & 25.2 & 68.4 & 33.3 \\
         Gemma-3n E4B & 57.3 & 5.6 & 54.5 & 8.2 & 73.7 & 16.7 & 75.1 & 31.8 & 56.6 & 22.1 & 53.1 & 29.3 \\
         Qwen2.5 7B Turbo& 55.7 & 7.2 & 52.4 & 10.6 & 66.9 & 10.5 & 57.9 & 19.4 & 58.8 & 27.1 & 59.6 & 34.8 \\
         Llama-3.3 70B Lite & 62.4 & 8.6 & 52.6 & 7.5 & 72.3 & 16.1 & 65.5 & 24.3 & 55.2 & 28.8 & 47.2 & 27.2 \\
         Llama-3 8B Lite& 34.8 & 6.8 & 35.0 & 4.6 & 46.7 & 16.6 & 49.8 & 19.7 & 36.1 & 21.1 & 44.9 & 23.5 \\
         \bottomrule
    \end{tabular}
    }
    \caption{The average satisfaction rate from GPT-5 Nano. All values are reported as percentages. The maximum approximate standard error is 5.1 percentage points.}
    \label{tb:main_gpt}
\end{table*}

\begin{table}
    \centering
    \begin{tabular}{ccc}
    \toprule
          \multicolumn{3}{c}{Common}    \\
    \midrule
        News & Story & Blog \\
        53.4$_{\pm4.2}$ &  60.5$_{\pm2.7}$ & 54.1$_{\pm2.9}$\\
        \midrule
         \multicolumn{3}{c}{Single} \\
         \midrule
        News & Story & Blog \\
        23.4$_{\pm2.2}$ & 30.1$_{\pm2.5}$ & 18.9$_{\pm2.3}$ \\
        \bottomrule
    \end{tabular}
    \caption{Percentages of constraints that are satisfied by the base article and thus, revised by our methods. }
    \label{tb:revision_perc}
\end{table}

\begin{table*}[t]
    \centering
    \scalebox{0.85}{
    \begin{tabular}{c|cccc|cccc|cccc}
        \toprule
         & \multicolumn{4}{c|}{News} & \multicolumn{4}{c|}{Story} & \multicolumn{4}{c}{Blog} \\
         & \multicolumn{2}{c}{Single} & \multicolumn{2}{c|}{Common} & \multicolumn{2}{c}{Single} & \multicolumn{2}{c|}{Common} & \multicolumn{2}{c}{Single} & \multicolumn{2}{c}{Common} \\
         & Org & Core  & Org & Core  &  Org & Core & Org & Core  & Org & Core  &  Org & Core   \\
         \midrule
         %Base Article & 0.562 & - & 0.544 & - & 0.483 & - & 0.428 & - & 0.578 & - & 0.569 & - \\
         Claude Sonnet 4.5 & 0.616 & 0.526 & 0.577 & 0.500 & 0.479 & 0.440 & 0.388 & 0.359 & 0.655 & 0.576 & 0.604 & 0.561 \\
         Claude Haiku 4.5 & 0.598 & 0.522 & 0.565 & 0.503 & 0.472 & 0.433 & 0.395 & 0.358 & 0.644 & 0.586 & 0.601 & 0.569 \\
         GPT-5 Mini & 0.598 & 0.509 & 0.561 & 0.485 & 0.482 & 0.431 & 0.405 & 0.357 & 0.628 & 0.579 & 0.607 & 0.570 \\
         GPT-5 Nano & 0.609 & 0.533 & 0.567 & 0.504 & 0.514 & 0.453 & 0.446 & 0.401 & 0.643 & 0.579 & 0.608 & 0.571 \\
         LFM2 24B A2B& 0.577 & 0.518 & 0.560 & 0.506 & 0.481 & 0.446 & 0.433 & 0.386 & 0.602 & 0.573 & 0.579 & 0.566 \\
         Gemma-3n E4B & 0.610 & 0.525 & 0.580 & 0.508 & 0.484 & 0.430 & 0.402 & 0.360 & 0.631 & 0.572 & 0.604 & 0.567 \\
         Qwen2.5 7B Turbo& 0.649 & 0.534 & 0.617 & 0.516 & 0.536 & 0.444 & 0.420 & 0.371 & 0.677 & 0.595 & 0.640 & 0.581 \\
         Llama-3.3 70B Lite & 0.653 & 0.544 & 0.608 & 0.518 & 0.524 & 0.451 & 0.424 & 0.371 & 0.668 & 0.584 & 0.626 & 0.574 \\
         Llama-3 8B Lite& 0.617 & 0.512 & 0.608 & 0.509 & 0.504 & 0.412 & 0.447 & 0.351 & 0.644 & 0.587 & 0.635 & 0.580 \\
         \bottomrule
    \end{tabular}
    }
    \caption{The average similarity between each constraint to the closest sentence in the response. Maximum SE across all cells is 0.023.}
    \label{tb:main_sim}
\end{table*}

\section{Examples of Synthesized Constraints}
To understand how our methods alleviate the copy-and-paste problem, we present constraints from one instance in each domain.

Table~\ref{tab:qual_copy} shows the copying problem in the news domain, where the reference article is about student walkouts in Alaska. The reference article
mentions a specific organization by name. When the constraint is synthesized
from this single article, the same name is copied into the constraint, so the
evaluated LLM only needs to repeat the name to satisfy it. When the constraint
is synthesized from two similar articles, the name disappears and the
constraint asks for a type of group instead, so the LLM has to decide which
groups to write about.

Table~\ref{tab:qual_copy_blog} shows the same pattern in the blog domain.
The reference blog explains how to add images on a specific publishing
platform. When the constraint is synthesized from this single blog, the
platform name and its interface are copied into the constraint, so the
evaluated LLM only needs to repeat them to satisfy it. When the constraint
is synthesized from two similar blogs, the platform name disappears and the
constraint asks for methods of adding images to a website in general, so the
LLM has to decide which platform and procedure to write about.

Table~\ref{tab:qual_copy_story} shows the same problem in the story domain.
The reference story describes objects breaking during an outburst, and the
single-article constraint carries those exact objects into the instruction. The
common constraint from two stories asks only for a setting to be established,
so the model must choose the objects and the scene itself.

\begin{table}[H]
\centering
\small
\setlength{\tabcolsep}{4pt}
\renewcommand{\arraystretch}{1.2}
\begin{tabular}{@{}>{\raggedright\arraybackslash}p{0.20\columnwidth}>{\raggedright\arraybackslash}p{0.74\columnwidth}@{}}
\toprule
\textbf{Source} & \textbf{Text} \\
\midrule
Reference Article A & ``\ldots The demonstrations were organized by the \textbf{Alaska Association of Student Government}.'' \\
\addlinespace
Single Constraint & ``Identify the \textbf{Alaska Association of Student Government} as the organizer of the protest.'' \\
\addlinespace
Common Constraint & ``Describe the organizational efforts behind the protests, including the involvement of \textbf{student government or community groups}.'' \\
\bottomrule
\end{tabular}
\caption{The copying problem in the news domain. The constraint from a single article copies a name from the reference
article. The common constraint from two articles keeps the same idea without
the name. Bold marks the copied text and its general version.}
\label{tab:qual_copy}
\end{table}

\begin{table}[H]
\centering
\small
\setlength{\tabcolsep}{4pt}
\renewcommand{\arraystretch}{1.2}
\begin{tabular}{@{}>{\raggedright\arraybackslash}p{0.20\columnwidth}>{\raggedright\arraybackslash}p{0.74\columnwidth}@{}}
\toprule
\textbf{Source} & \textbf{Text} \\
\midrule
Reference Blog A & ``Adding images in a \textbf{WordPress} powered site is pretty easy, as they can be stored and managed in a single place i.e.\ \textbf{Media Library}.'' \\
\addlinespace
Single Constraint & ``Explain the step-by-step process of uploading images directly to the \textbf{Media Library} through the \textbf{WordPress} dashboard.'' \\
\addlinespace
Common Constraint & ``Explain different methods of adding images to \textbf{a website} for better content presentation.'' \\
\bottomrule
\end{tabular}
\caption{The copying problem in the blog domain.}
\label{tab:qual_copy_blog}
\end{table}

\begin{table}[H]
\centering
\small
\setlength{\tabcolsep}{4pt}
\renewcommand{\arraystretch}{1.2}
\begin{tabular}{@{}>{\raggedright\arraybackslash}p{0.20\columnwidth}>{\raggedright\arraybackslash}p{0.74\columnwidth}@{}}
\toprule
\textbf{Source} & \textbf{Text} \\
\midrule
Reference Story A & ``\textbf{Fine China and scented candles} clattered the floor...'' \\
\addlinespace
Single Constraint & ``Incorporate detailed imagery of the prince's chambers, including \textbf{broken fine china and scented candles} after the king's outburst.'' \\
\addlinespace
Common Constraint & ``Begin by establishing the \textbf{setting or environment} where the main events will unfold.'' \\
\bottomrule
\end{tabular}
\caption{The copying problem in the story domain.}
\label{tab:qual_copy_story}
\end{table}

% Table~\ref{tab:qual_revision} shows how the revision methods affect the constraints.
% \underline{Direct} makes it harder by adding a very specific situation that a real user would rarely ask for.
% \underline{Eval} makes it harder by asking for more information, such as which
% groups were involved and what they did, without telling the LLM what should
% happen in the article. Neither revision brings back a name from the reference
% article.

% \begin{table}[H]
% \centering
% \small
% \setlength{\tabcolsep}{4pt}
% \renewcommand{\arraystretch}{1.2}
% \begin{tabular}{@{}>{\raggedright\arraybackslash}p{0.20\columnwidth}>{\raggedright\arraybackslash}p{0.74\columnwidth}@{}}
% \toprule
% \textbf{Method} & \textbf{Constraint} \\
% \midrule
% Common + None & ``Include detailed accounts of the middle portion of events during the protests or walkouts led by students.'' \\
% \addlinespace
% Common + Direct & ``Present detailed accounts of the mid-protest organizational challenges, such as coordinating across diverse student groups with conflicting priorities.'' \\
% \addlinespace
% Common + Eval & ``Detail the organizational logistics behind the protests, naming involved student groups or community coalitions and their roles.'' \\
% \bottomrule
% \end{tabular}
% \caption{The constraints after each revision method.}
% \label{tab:qual_revision}
% \end{table}

\section{Method Details}
\label{sec:method_details}

We present all the prompts we used in the appendix. The story constraint synthesizing prompt is the same as the prompt for blog except replacing the blog category name with story. We found that \textit{GPT-4.1 Mini} often generates longer articles than the recommended length, so we ask \textit{GPT-4.1 Mini} to generate base article using $400$ words in Template~\ref{prompt:base} to keep the average length of base articles to be around $500$. 

%ask the model to shuffle the order of the constraints
%sbert uses cosine similarity

% ----------------------------- 1 -----------------------------
% Constraint Generation - Blog  (prompts.py) | variables: none
\begin{myprop}
\begin{lstlisting}
Constraint Generation - Blog

You are a writing expert. I am going to give you a blog as an input.
You can assume that a large language model (LLM) generated the blog.

Your task has two parts:
1. Identify the main task of the blog in one sentence.
   - For example: "The main task is to write a blog about strategies for successful remote working."
   - Phrase the main task as an instruction.
2. Generate a set of 39 free-form constraints that you think might have been given to the LLM to generate the blog.
   - DO NOT REPEAT CONSTRAINTS.
   - Constraints must be atomic (a single indivisible condition). If a constraint can be broken into smaller constraints, do so.
   - Avoid proper nouns in your constraints.
   - Constraints should drive at least a few sentences in the blog (do not write constraints that map to only one line).
   - Constraints must strictly pertain to the content, ideas, arguments, or narrative direction of the blog and should influence how the blog develops.
   - If (and only if) you cannot write 39 atomic, content-based constraints, give stylistic constraints based on how the blog is written (tone, use of examples, formatting, etc.).
   - Write all constraints in the form of instructions. For example: "The blog should include practical tips."
   - CRITICAL RANDOMIZATION STEP: You must disrupt the chronological flow. To do this, strictly follow this pattern: Write your first constraint about the conclusion of the blog. Write your second constraint about the introduction. Write your third constraint about the middle. Continue jumping back and forth across the timeline of the narrative for all 39 constraints. The final list must feel completely scrambled with no narrative arc.

Here is a worked example to guide you:

Input Blog:
Working from home has become the new normal for millions of professionals worldwide. While it offers flexibility and eliminates commutes, it also presents unique challenges that can impact both productivity and well-being.

To optimize your home workspace, start by creating a dedicated area free from distractions. This space should have good lighting, comfortable seating, and all necessary equipment within reach. Many experts recommend facing a window for natural light, which can boost mood and energy levels.

Establish clear boundaries between work and personal time. Set specific work hours and stick to them, just as you would in an office. Communicate these boundaries to family members or housemates to minimize interruptions during work hours.

Take regular breaks throughout the day. The Pomodoro Technique, which involves 25-minute focused work sessions followed by 5-minute breaks, can help maintain concentration and prevent burnout. Use break time to stretch, hydrate, or take a short walk.

Stay connected with colleagues through regular video calls and instant messaging. This helps maintain team cohesion and prevents feelings of isolation. Schedule virtual coffee breaks or team-building activities to foster relationships.

Finally, prioritize your physical and mental health. Maintain a regular exercise routine, eat nutritious meals, and get adequate sleep. Consider meditation or mindfulness practices to manage stress and maintain focus.

Output:
Main Task: Write a blog about strategies for successful remote working.

Constraints:
1. Require the setting of defined working hours.
2. Explain the risks of isolation if connection practices are neglected.
3. Warn about the risk of burnout without intentional self-care.
4. Explain how the removal of commuting affects time use and daily rhythm.
5. Suggest strategies for maintaining healthy eating while at home.
6. Emphasize reducing environmental distractions in that space.
7. Argue for the necessity of regular breaks during the workday.
8. Encourage informal online gatherings to maintain rapport.
9. Recommend mindfulness or meditation as stress-management tools.
10. Stress the value of adhering consistently to those hours.
11. Show how workspace ergonomics (chair, desk) influence long-term health.
12. Conclude with a call to action urging readers to adopt concrete changes immediately.
13. Link exercise directly to improved cognitive performance and focus.
14. Establish remote work as a global trend that has transformed professional life.
15. Connect emotional well-being to overall job performance and satisfaction.
16. Explain how balanced nutrition influences concentration and resilience.
17. Emphasize communicating work schedules to others in the household.
18. Identify productivity as a central theme in remote work discussions.
19. Recommend scheduled video calls to replicate face-to-face connection.
20. Recommend creating a physically separate space for work at home.
21. Stress the need for essential tools and equipment to be easily accessible.
22. Recommend light physical movement or stretching during pauses.
23. Integrate workspace, scheduling, health, and social practices into a unified remote-work strategy.
24. Stress the importance of sleep in sustaining energy and productivity.
25. Highlight the role of hydration and snacks in sustaining energy across breaks.
26. Suggest environmental cues (like decor or layout) that reinforce the sense of a work zone.
27. Describe the importance of adequate lighting for focus and energy.
28. Introduce one structured time-management method, such as work intervals.
29. Show how mental health practices support long-term work sustainability.
30. Show how shared rituals (e.g., virtual coffee breaks) strengthen belonging.
31. Highlight well-being as equally important alongside productivity.
32. Warn about the risk of personal time erosion without such boundaries.
33. Contrast the flexibility of remote work with the new challenges it creates.
34. Suggest instant messaging as a tool for quick, ongoing collaboration.
35. Highlight how a clear boundary between workspace and leisure areas aids focus.
36. Show how enforcing those boundaries prevents interruptions.
37. Recommend establishing a routine for daily physical exercise.
38. Explain how breaks counteract mental fatigue and sustain performance.
39. Stress that remote work requires deliberate maintenance of social contact.

Now use the same approach for the next input blog.
\end{lstlisting}
\end{myprop}

% ----------------------------- 2 -----------------------------
% Constraint Generation - News  (news_prompts.py) | variables: none
\begin{myprop}
\begin{lstlisting}
Constraint Generation - News

You are a writing expert. I am going to give you a news article as an input.
You can assume that a large language model (LLM) generated the news article.

Your task has two parts:
1. Identify the main task of the news article in one sentence.
   - For example: "The main task is to write a news article about 43rd Annual Greek Experience Festival June 7-9 In Danbury."
   - Phrase the main task as an instruction.
2. Generate a set of 39 free-form constraints that you think might have been given to the LLM to generate the news article.
   - DO NOT REPEAT CONSTRAINTS.
   - Constraints must be atomic (a single indivisible condition). If a constraint can be broken into smaller constraints, do so.
   - Avoid proper nouns in your constraints.
   - Constraints should drive at least a few sentences in the news article (do not write constraints that map to only one line).
   - Constraints must strictly pertain to the content, ideas, arguments, or narrative direction of the news article and should influence how the news article develops.
   - If (and only if) you cannot write 39 atomic, content-based constraints, give stylistic constraints based on how the news article is written (tone, use of examples, formatting, etc.).
   - Write all constraints in the form of instructions. For example: "The story should center on a disappearance." "Authorities should coordinate a large-scale search effort."
   - Do not write constraints in the same order or phrasing as the article text. Randomize the order of the constraints.

Return the result ONLY in this exact format, with no extra text, commentary, or Markdown:

Main Task: <one sentence instruction>

Constraints:
1. <constraint 1>
2. <constraint 2>
...
39. <constraint 39>

Here is a worked example to guide you:

Input:
Share event 43rd Annual Greek Festival Assumption Greek Orthodox Church in Danbury will host its 43rd Annual Greek Festival celebrating the rich culture of our Greek heritage June 7, 8 and 9 on the Church grounds - Clapboard Ridge Road. Food and Drink, Live Music, & Folk Dancing! Come enjoy our warm hospitality, learn more about our beautiful Byzantine Church, and experience our Orthodox Christian traditions. Admission is free to this fun-filled celebration that showcases many of the most beloved aspects of Greek culture including Folk Dancers performing in authentic costumes, live music, artisan crafts, wine and of course - delicious food. It's easy to find! Just take exit 5 off of I-84 and follow the signs. There is additional free parking and shuttle service available at Danbury High School, as well.

Output:
Main Task: Write a news article about 43rd Annual Greek Experience Festival June 7-9 In Danbury.

Constraints:
1. The story should center on an upcoming cultural celebration.
2. The event should be hosted by a community or religious organization.
3. The organization should have a long-standing tradition of holding this event annually.
4. The event should celebrate a specific cultural or ethnic heritage.
5. The story should indicate that the celebration spans multiple consecutive days.
6. The event's location should be mentioned with reference to a recognizable area or landmark.
7. The story should state that the event is held on the organization's premises.
8. The festival should promote awareness of traditional customs and community values.
9. The story should highlight the preservation of cultural identity as a theme.
10. The event should emphasize the warm hospitality of the hosting community.
11. The narrative should invite the public to experience the featured culture firsthand.
12. Religious or spiritual elements should be included to connect the event to its heritage.
13. The article should present the event as both entertaining and educational.
14. The story should mention a variety of attractions or activities.
15. Traditional food and drink should be featured prominently.
16. The festival should include live performances for entertainment.
17. Folk dancing should be identified as a key cultural activity.
18. Authentic costumes should be highlighted as part of the performances.
19. Live music should be described as enhancing the festive atmosphere.
20. Artisan crafts should be included as part of the offerings.
21. Visitors should have opportunities to learn about cultural traditions.
22. The tone should convey warmth, joy, and community spirit.
23. The story should encourage public participation and inclusivity.
24. The article should express pride in the cultural heritage being celebrated.
25. The narrative should appeal to the senses through vivid descriptions of the experience.
26. The story should state that admission is free.
27. Practical details for attendees should be provided.
28. Directions or reference points should be given for ease of navigation.
29. The article should mention additional parking or transportation options.
30. Accessibility and convenience should be emphasized to attract attendees.
31. The story should begin with a clear announcement of the event and its host.
32. A section should describe the cultural or historical significance of the event.
33. The middle part should list the main attractions and entertainment.
34. The closing section should include logistical information for visitors.
35. The tone throughout should remain promotional and welcoming.
36. The structure should progress from general overview to detailed information.
37. The story should use present or future tense for immediacy.
38. The language should be simple, lively, and community-oriented.
39. The narrative should maintain an informative tone typical of local event announcements.

Now use the same approach for the next input news article.
\end{lstlisting}
\end{myprop}

% ----------------------------- 3 -----------------------------
% Common Constraint Generation - Blog  (prompts.py) | variables: {blog1} {blog2}
\begin{myprop}
\begin{lstlisting}
Common Constraint Generation - Blog

You are a writing expert. You will be given two blogs (Blog A and Blog B) as input.
You can assume that a large language model (LLM) generated each blog.

Your task has two parts:
1. Identify a common main task that applies to BOTH blogs.
   - The main task must be implied by BOTH Blog A and Blog B - every detail in the task must be directly present in both blogs.
   - The more similar the blogs, the more specific the main task can be.
   - The more dissimilar the blogs, the more general the main task should be.
   - Phrase the main task as an instruction. Example: "Write a blog about strategies for successful remote working."

2. Generate a set of 39 free-form constraints that apply to BOTH blogs.
   - Each constraint must be satisfied by BOTH Blog A and Blog B.
   - DO NOT REPEAT CONSTRAINTS.
   - Constraints must be atomic (a single indivisible condition). If a constraint can be broken into smaller constraints, do so.
   - Avoid proper nouns in your constraints.
   - Constraints should drive at least a few sentences in both blogs (do not write constraints that map to only one line).
   - Constraints must strictly pertain to the content, ideas, arguments, or narrative direction and should influence how the blogs develop.
   - The more similar the blogs, the more specific and detailed the constraints should be.
   - The more dissimilar the blogs, the more general and abstract the constraints should be (e.g., "Include a conclusion" or "Use examples").
   - If the blogs are so dissimilar that you cannot find 39 content-based constraints, give stylistic constraints based on how both blogs are written (tone, use of examples, formatting, etc.).
   - Write all constraints in the form of instructions. Example: "The blog should include practical tips."
   - CRITICAL RANDOMIZATION STEP: Disrupt chronological flow. Write your first constraint about conclusions, second about introductions, third about middles. Continue jumping across the timeline. The final list must feel completely scrambled.

Here are examples showing how similarity affects output:

Example 1 (Dissimilar Blogs):
Blog A: A technical blog about Python web scraping with BeautifulSoup, including code examples and CSV export.
Blog B: A narrative blog about traveling through the Amazon rainforest, focusing on wildlife encounters.

Main Task: Write a blog that provides detailed information on a topic.

Constraints:
1. End with a summary or concluding thought.
2. Open with an introduction that establishes the topic.
3. Include specific examples to illustrate points.
4. Maintain a clear and organized structure.
5. Use descriptive language to engage the reader.
... (remaining constraints would be similarly general)

Example 2 (Very Similar Blogs):
Blog A: A blog about remote work productivity, discussing workspace setup, time management with Pomodoro, and work-life boundaries.
Blog B: A blog about working from home effectively, covering dedicated workspace creation, scheduled breaks, and separating work from personal time.

Main Task: Write a blog about strategies for successful remote working.

Constraints:
1. Conclude with a call to action urging readers to implement changes.
2. Establish remote work as a significant trend in professional life.
3. Recommend creating a dedicated workspace at home.
4. Emphasize reducing distractions in the work environment.
5. Argue for the importance of regular breaks during work.
6. Recommend a structured time-management approach.
7. Stress setting clear boundaries between work and personal time.
8. Suggest communicating work schedules to household members.
... (remaining constraints would be similarly specific)

Now apply this approach to the following two blogs:

Blog A:
{blog1}

Blog B:
{blog2}

Output Format:
Main Task: [one sentence instruction]

Constraints:
1. [constraint]
2. [constraint]
...
39. [constraint]
\end{lstlisting}
\end{myprop}

% ----------------------------- 4 -----------------------------
% Common Constraint Generation - News  (news_prompts.py) | variables: {blog1} {blog2}
\begin{myprop}
\begin{lstlisting}
Common Constraint Generation - News

You are a writing expert. You will be given two news articles (Article A and Article B) as input.
You can assume that a large language model (LLM) generated each news article.

Your task has two parts:
1. Identify a common main task that applies to BOTH news articles.
   - The main task must be implied by BOTH news article A and news article B - every detail in the task must be directly present in both the news articles.
   - The more similar the news articles, the more specific the main task can be.
   - The more dissimilar the news articles, the more general the main task should be.
   - Phrase the main task as an instruction. Example: "Report on recent developments related to changes in workplace policies."

2. Generate a set of 39 free-form constraints that apply to BOTH articles.
   - Each constraint must be satisfied by article A and article B.
   - DO NOT REPEAT CONSTRAINTS.
   - Constraints must be atomic (a single indivisible condition). If a constraint can be broken into smaller constraints, do so.
   - Avoid proper nouns in your constraints.
   - Constraints should drive at least a few sentences in both the articles (do not write constraints that map to only one line).
   - Constraints must strictly pertain to the content, ideas, arguments, or narrative direction and should influence how the articles develop.
   - The more similar the articles, the more specific and detailed the constraints should be.
   - The more dissimilar the articles, the more general and abstract the constraints should be (e.g., "Include a conclusion" or "Use examples").
   - If the articles are so dissimilar that you cannot find 39 content-based constraints, give stylistic constraints based on how both articles are written (tone, use of examples, formatting, etc.).
   - Write all constraints in the form of instructions. Example: "The articles should include practical tips."
   - CRITICAL RANDOMIZATION STEP: Disrupt chronological flow. Write your first constraint about conclusions, second about introductions, third about middles. Continue jumping across the timeline. The final list must feel completely scrambled.

Here are examples showing how similarity affects output:

Example 1 (Dissimilar Articles):
Article A: A breaking news report about a natural disaster focusing on immediate impacts and emergency response.
Article B: An investigative article examining long-term issues in public infrastructure funding.

Main Task: Write a news article that informs readers about a significant real-world issue.

Constraints:
1. End with information that looks ahead to future developments.
2. Open with a clear lead summarizing the core issue.
3. Provide factual details supported by evidence.
4. Maintain a neutral and objective reporting tone.
5. Include contextual background to help readers understand the issue.
... (remaining constraints would be similarly general)

Example 2 (Very Similar Articles):
Article A: A news article reporting on the rise of remote work, covering recent survey findings, employer policy changes, and statements from workers about productivity and work-life balance.
Article B: A news article reporting on the continued expansion of work-from-home arrangements, highlighting company guidelines, expert commentary on productivity, and employee experiences with managing work-life boundaries.

Main Task: Report on recent trends in remote work and how they affect productivity and work-life balance.

Constraints:
1. Conclude by discussing how remote work policies may evolve in the near future.
2. Open with a lead that summarizes the growing prevalence of remote work.
3. Attribute productivity claims to surveys, studies, or expert commentary.
4. Include perspectives from workers experiencing remote work firsthand.
5. Report on employer or organizational policy changes related to remote work.
6. Explain challenges related to managing work-life boundaries.
7. Provide background on how remote work practices have changed over time.
8. Maintain a neutral, factual, and non-prescriptive tone throughout.
... (remaining constraints would be similarly specific)

Now apply this approach to the following two news articles:

Article A:
{blog1}

Article B:
{blog2}

Output Format:
Main Task: [one sentence instruction]

Constraints:
1. [constraint]
2. [constraint]
...
39. [constraint]
\end{lstlisting}
\end{myprop}

% ----------------------------- 5 -----------------------------
% Base Generation - Shared  (prompts.py / news_prompts.py, identical) | variables: {content_type}
\begin{myprop}
\begin{lstlisting}
Base Generation 

You are a creative writing expert. I will give you a task description, and you need to generate {content_type} content that fulfills the task.

The content should be:
- Well-structured and coherent
- Engaging and creative
- Of appropriate length (aim for 400 words)
- Natural and authentic in tone

Generate the {content_type} based on the following task:
\end{lstlisting}
\label{prompt:base}
\end{myprop}

% ----------------------------- 6 -----------------------------
% Base Revision  (identical) | variables: {content_type} {task} {base_content} {constraints}
\begin{myprop}
\begin{lstlisting}
Base Revision

You are a creative writing expert. I will give you:
1. A task description
2. Base {content_type} content
3. A list of 39 constraints

Your job is to revise and expand the base content to satisfy the constraints while maintaining coherence, quality, and natural flow. Restrict the length of output to 500 words.

Instructions:
- Keep the core ideas from the base content
- Integrate constraints seamlessly where they fit naturally
- Prioritize natural flow and readability over satisfying every single constraint
- It is acceptable to skip constraints if they force the writing to be awkward
- Maintain a natural, engaging writing style
- Ensure the content flows logically
- Do not mention the constraints explicitly in the content
- Aim for completeness - the content should feel finished and polished

Task: {task}

Base Content:
{base_content}

Constraints to satisfy:
{constraints}

Generate the revised {content_type} that satisfies all constraints:
\end{lstlisting}
\end{myprop}

% ----------------------------- 8 -----------------------------
% Direct Generation  (prompts.py) | variables: {content_type} {task} {constraints}
\begin{myprop}
\begin{lstlisting}
Direct Generation

You are a creative writing expert. I will give you:
1. A task description
2. A list of constraints to satisfy

Your job is to generate a complete, high-quality {content_type} that fulfills the task and satisfies as many constraints as possible while maintaining natural flow and coherence. Restrict the length of output to 500 words.

Instructions:
- Generate engaging, creative content that directly addresses the task
- Integrate constraints seamlessly where they fit naturally
- Prioritize natural flow, readability, and quality over satisfying every single constraint
- Maintain a natural, engaging writing style
- Ensure the content flows logically and feels complete
- Do not mention the constraints explicitly in the content
- The final output should feel polished and professional

Task: {task}

Constraints to satisfy:
{constraints}

Generate the {content_type}:
\end{lstlisting}
\end{myprop}

% ----------------------------- 9 -----------------------------
% Evaluation (prompts.py, active) | variables: {content_type_capitalized} {content} {constraints}
\begin{myprop}
\begin{lstlisting}
Evaluation

You are a strict constraint evaluator.

You will be given:
1. Content (a story or blog)
2. A numbered list of constraints

Your task:
Evaluate each constraint independently and determine whether it is fully satisfied by the content.

CRITICAL INDEX RULE (MANDATORY):
Each output line i MUST evaluate ONLY constraint i.
Do not merge, skip, combine, or split constraints.
Even if multiple constraints appear similar, each MUST receive its own line.

OUTPUT RULES (MANDATORY):
1. Output EXACTLY one line per constraint, in numeric order, followed by ONE final line.
2. Do NOT output any extra text, headers, explanations, summaries, or blank lines.
3. Each constraint line MUST follow one of these exact formats:

   i. Yes - "<exact sentence or excerpt from the content that proves the constraint>"
   OR
   i. No - <one short sentence explaining why the constraint is not satisfied>

4. Mark "Yes" ONLY if the constraint is explicitly and completely satisfied.
   - Partial satisfaction = No
   - Inference, implication, or interpretation = No
   - If uncertain, mark No

EVALUATION RULES:
- Semantic constraints must be clearly stated in the content.
- If you must explain or justify a Yes beyond quoting text, mark No.
- Do not stretch meanings or combine multiple moments to satisfy one constraint.

WORD COUNT RULES:
- "less than N words" -> Yes only if word_count < N
- "about N words" or "approximately N words" -> Yes only if word_count is within +/-10 percent of N

FINAL COUNT:
After evaluating all constraints, output exactly:
Number of constraints satisfied: X

X MUST equal the total number of lines marked "Yes" above.
Silently recount before printing X.

Now evaluate:

{content_type_capitalized}:
{content}

Constraints:
{constraints}

Output:
\end{lstlisting}
\end{myprop}

% ----------------------------- 11 -----------------------------
% Evaluation - Terse  (prompts.py) | variables: {content_type_capitalized} {content} {constraints}
\begin{myprop}
\begin{lstlisting}
Evaluation - Yes/No config

You are a strict constraint evaluator.

You will be given:
1. Content (a story or blog)
2. A numbered list of constraints

Your task:
Evaluate each constraint independently and determine whether it is fully satisfied by the content.

CRITICAL INDEX RULE (MANDATORY):
Each output line i MUST evaluate ONLY constraint i.
Do not merge, skip, combine, or split constraints.
Even if multiple constraints appear similar, each MUST receive its own line.

OUTPUT RULES (MANDATORY):
1. Output EXACTLY one line per constraint, in numeric order.
2. Each line MUST be EXACTLY one of these two formats, with NOTHING else:
       i. Yes
       i. No
3. Do NOT output quotes, reasons, explanations, headers, summaries, or blank lines.
4. Mark "Yes" ONLY if the constraint is explicitly and completely satisfied.
   - Partial satisfaction = No
   - Inference, implication, or interpretation = No
   - If uncertain, mark No

EVALUATION RULES:
- Semantic constraints must be clearly stated in the content.
- Do not stretch meanings or combine multiple moments to satisfy one constraint.

WORD COUNT RULES:
- "less than N words" -> Yes only if word_count < N
- "about N words" or "approximately N words" -> Yes only if word_count is within +/-10 percent of N

FINAL COUNT:
After all constraint lines, output exactly:
Number of constraints satisfied: X
X MUST equal the number of lines marked "Yes" above. Silently recount before printing X.

Now evaluate:

{content_type_capitalized}:
{content}

Constraints:
{constraints}

Output:
\end{lstlisting}
\end{myprop}

% ----------------------------- 13 -----------------------------
% Summarization - Shared  (identical) | variables: {content_type} {target_pct}
\begin{myprop}
\begin{lstlisting}
Summarization 

Given the {content_type} post, rewrite a summarized version that is approximately {target_pct}% of the original length.
Output only the summarized {content_type}, with no preamble or explanation.

[Runtime appends, after the template above:]
{content_type_capitalized} to summarize:
{content}
\end{lstlisting}
\end{myprop}

% ----------------------------- 14 -----------------------------
% Constraint Replacement, difficulty-based (no eval) - Blog  (prompts.py) | variables: {main_task} {original_constraints} {base_content}
\begin{myprop}
\begin{lstlisting}
Constraint Replacement - direct

You are a writing expert. You are given:
1. A main task
2. A set of 39 constraints for that task
3. A base story/blog written for that task

Your job is to create a REVISED set of 39 constraints by replacing constraints that are TOO EASY with NEW, HARDER constraints that would be more difficult for a writer to satisfy.

Only Output the revised constraints, with no preamble or explanation.

Requirements for replacement constraints:
- Must be relevant to the same main task
- Should be harder to satisfy than the ones they replace
- Should require significant modification to satisfy
- Must be atomic (single indivisible condition)
- Must be content-based (not just stylistic)
- Should maintain overall coherence with kept constraints

Instructions:
1. Keep the main task EXACTLY the same
2. Identify constraints that seem easy or generic and replace them with harder, more specific ones
3. Keep constraints that are already sufficiently difficult
4. Ensure you still have exactly 39 constraints total
5. Number the revised constraints 1-39
6. Randomize the order (don't put all new constraints at the end)

Output Format:
Main Task: [same as original]

Revised Constraints:
1. [constraint - either kept from original or new replacement]
2. [constraint]
...
39. [constraint]

---

Input:

Main Task: {main_task}

Original Constraints:
{original_constraints}

Base Content:
{base_content}

Output:
\end{lstlisting}
\end{myprop}

% ----------------------------- 15 -----------------------------
% Constraint Replacement with eval - Blog  (prompts.py) | variables: {main_task} {original_constraints} {base_content} {satisfaction_results}
\begin{myprop}
\begin{lstlisting}
Constraint Replacement - Eval

You are a writing expert. You are given:
1. A main task
2. A set of 39 constraints for that task
3. A base story/blog written for that task
4. Evaluation results showing which constraints are already satisfied by the base content

Your job is to create a REVISED set of 39 constraints by replacing the constraints that are already satisfied (marked "Yes") with NEW, HARDER constraints that are NOT satisfied by the base content.

Only Output the revised constraints, with no preamble or explanation.

Requirements for replacement constraints:
- Must be relevant to the same main task
- Must NOT be satisfied by the current base content
- Should require significant modification to satisfy
- Must be atomic (single indivisible condition)
- Must be content-based (not just stylistic)
- Should maintain overall coherence with kept constraints

Instructions:
1. Keep the main task EXACTLY the same
2. Keep all constraints marked "No" (not satisfied) unchanged
3. Replace all constraints marked "Yes" (satisfied) with new, harder constraints
4. Ensure you still have exactly 39 constraints total
5. Number the revised constraints 1-39
6. Randomize the order (don't put all new constraints at the end)

Output Format:
Main Task: [same as original]

Revised Constraints:
1. [constraint - either kept from original or new replacement]
2. [constraint]
...
39. [constraint]

---

Input:

Main Task: {main_task}

Original Constraints:
{original_constraints}

Base Content:
{base_content}

Evaluation Results (Yes = satisfied, No = not satisfied):
{satisfaction_results}

Output:
\end{lstlisting}
\end{myprop}

% ----------------------------- 17 -----------------------------
% Pairwise Quality  (prompts.py) | variables: {content_a} {content_b}
\begin{myprop}
\begin{lstlisting}
Pairwise Quality

You are an expert writing evaluator. You will be given two pieces of content (Content A and Content B) and you must compare them on two metrics:

1. **Coherence**: Which content has better logical flow and cohesion?
2. **Likability**: Which content is more enjoyable and engaging to read?

For each metric:
- First, identify concrete issues for both contents
- Then provide a score out of 5 for both A and B
- Specify which content you prefer (A or B)
- Give a brief one-line reasoning for your preference

After evaluating both metrics, assign an overall winner (A or B) based on the category wins.

IMPORTANT: Follow the exact format shown in the example below. Do not add extra text.
Use PLAIN TEXT only - do NOT use any markdown formatting (no **bold**, no ## headers,
no bullet points or code fences). Reproduce the exact line format from the example,
e.g. `A - 2/5`, `B - 4/5`, `Preference: B`, and a final `Overall Winner: B` line with
the letter on the same line as the colon.

---

**Example:**

Content A:
Sarah walked through the forest. She arrived at the castle and confronted the villain. "You won't get away with this," she said. The villain laughed. Sarah defeated him and went home.

Content B:
Sarah walked through the forest, her heart pounding. Hours later, she found herself at the castle gates. She entered the throne room. The villain was waiting. "I've been expecting you," he said with a smile. Sarah raised her sword. They fought for what felt like hours. Finally, Sarah emerged victorious and began her journey home, exhausted but triumphant.

Coherence:
Issues in A: The transition from forest to castle is abrupt; unclear why Sarah suddenly knows the villain's location or how she defeated him.
Issues in B: The timeline is confusing - "hours later" suggests travel time but then "what felt like hours" for the fight creates temporal ambiguity.
A - 2/5
B - 3/5
Preference: B - Despite timeline issues, the narrative flow is more logical with clearer scene transitions.

Likability:
Issues in A: The dialogue feels stilted and generic; lacks descriptive detail and emotional depth.
Issues in B: The pacing drags slightly with phrases like "what felt like hours" being vague rather than engaging.
A - 2/5
B - 4/5
Preference: B - More engaging with vivid descriptions and emotional resonance despite minor pacing issues.

Overall Winner: B

---

Content A:
{content_a}

Content B:
{content_b}
\end{lstlisting}
\end{myprop}

% ----------------------------- 18 -----------------------------
% Naturalness Evaluation  (prompts.py) | variables: none in body (prompt/constraints appended at runtime)
\begin{myprop}
\begin{lstlisting}
Naturalness Evaluation

In this task, we want to understand how natural and appropriate different writing constraints feel when responding to given writing prompts. You will be shown a writing prompt and a pair of corresponding constraints. Each constraint describes something what could appear in a written response to their prompts.

What is a "constraint"?
A constraint is a requirement, guideline, or stylistic element that a writer might include when responding to the prompt. Some constraints may feel natural and helpful, while others may feel awkward, unnecessary, or overly restrictive.

Your Tasks
1. Preference Selection:
For each prompt, choose the constraint (A or B) that you would prefer to see included in a written response to their respective prompt.
2. Naturalness Scoring:
Independently score each constraint from 1 to 5 based on how likely you would include it in a natural, high-quality response to their respective prompt.
Score Scale Explanation

1 Never: You would almost never include this constraint; it feels very unnatural or inappropriate.
2 Rarely: You would include it only in unusual or forced situations.
3 Sometimes: You might include it depending on context, but it is not clearly necessary.
4 Often: You would usually include it; it feels natural and helpful.
5 Always: You would almost always include it; it feels essential and very natural.


Example
Prompt: Describe a challenge you faced while working on a team project.
Constraint A: Describe how communication among team members affected the outcome of the project.
Constraint B: Include what time of day the project was completed.
Constraint A feels natural and helpful because communication is often central when reflecting on team challenges. Constraint B is related to the project, but it is not essential to describing a specific challenge, making it less likely to appear in a typical response.

Important Notes
Preference and scores are related but not the same. You may prefer one constraint to be more natural over another even if both receive similar scores.

Please judge each constraint independently, not relative to the other constraint's score.
There are no "correct" answers. We care about your honest judgment.
Please read the prompt and each constraint carefully before answering.


Instructions
For each prompt, you will see two possible constraints (A and B).
Select which constraint (A or B) you think is more natural to include in a response to the prompt.

Score each constraint from 1 (never) to 5 (always) based on how likely you would include it in a natural, high-quality response.

Expected Output Format:
Preference: A
Score A: 4
Score B: 2

[Runtime appends, after the template above:]
Prompt: {prompt}
Constraint A: {constraint_a}
Constraint B: {constraint_b}
\end{lstlisting}
\end{myprop}

\section{Experiment Details}

%How do we segment the sentences?

We use sent\_tokenize in NLTK to segment the sentences in the generated article and in reference article to compute the similarity. In \Cref{tb:human_exp}, LLM coherence judge uses \textit{Claude Haiku 4.5} + \textit{GPT-5 Nano} as in our benchmark.

We extracted news articles from the dataset's preprocessed JSONL files, filtering to only include entries marked as valid news articles and retaining the title, content, publication date, and URL fields. We then removed articles with missing titles or content, stripped inline advertisement markers from the article body, and filtered by length, keeping only articles between 800 and 2,000 words.%\todo{to be confirmed}.

Finally, we restricted the blog and story candidate pool by length, keeping only blogs between 800 and 2,000 words to exclude fragments and unusually long documents.

% Mturk human task

% \begin{figure*}[h]
%     \centering
%     \includegraphics[width=0.8\textwidth]{figs/human_exp.pdf}
%     \caption{Mturk task template}
%     \label{fig:myfig}
% \end{figure*}

% \begin{figure*}[h]
%     \centering
%     \includegraphics[width=0.8\textwidth]{figs/user_constraints.pdf}
%     \caption{Mturk task template}
%     \label{fig:myfig}
% \end{figure*}

% \begin{figure*}[t]
%     \centering
%     \begin{minipage}[t]{0.48\textwidth}
%         \centering
%         \includegraphics[width=\textwidth]{figs/human_exp.pdf}
%         \caption{MTurk task template: naturalness \& preference scoring.}
%         \label{fig:human_exp}
%     \end{minipage}
%     \hfill
%     \begin{minipage}[t]{0.48\textwidth}
%         \centering
%         \includegraphics[width=\textwidth]{figs/user_constraints.pdf}
%         \caption{MTurk task template: user-written constraint reference.}
%         \label{fig:user_constraints}
%     \end{minipage}
% \end{figure*}

% \begin{figure*}[t]
%     \centering
%     \begin{minipage}[t]{0.48\textwidth}
%         \centering
%         \vtop{\hbox{\includegraphics[width=\textwidth]{figs/human_exp.pdf}}}
%         \caption{MTurk task template: naturalness \& preference scoring.}
%         \label{fig:human_exp}
%     \end{minipage}
%     \hfill
%     \begin{minipage}[t]{0.48\textwidth}
%         \centering
%         \vtop{\hbox{\includegraphics[width=\textwidth, trim=0 0 0 20pt, clip]{figs/user_constraints.pdf}}}
%         \caption{MTurk task template: user-written constraint reference.}
%         \label{fig:user_constraints}
%     \end{minipage}
    
% \end{figure*}

\subsection{MTurk Task Details and Templates}
\label{sec:appendix_mturk_templates}

\begin{figure}[htbp]
    \centering
    \includegraphics[width=\linewidth]{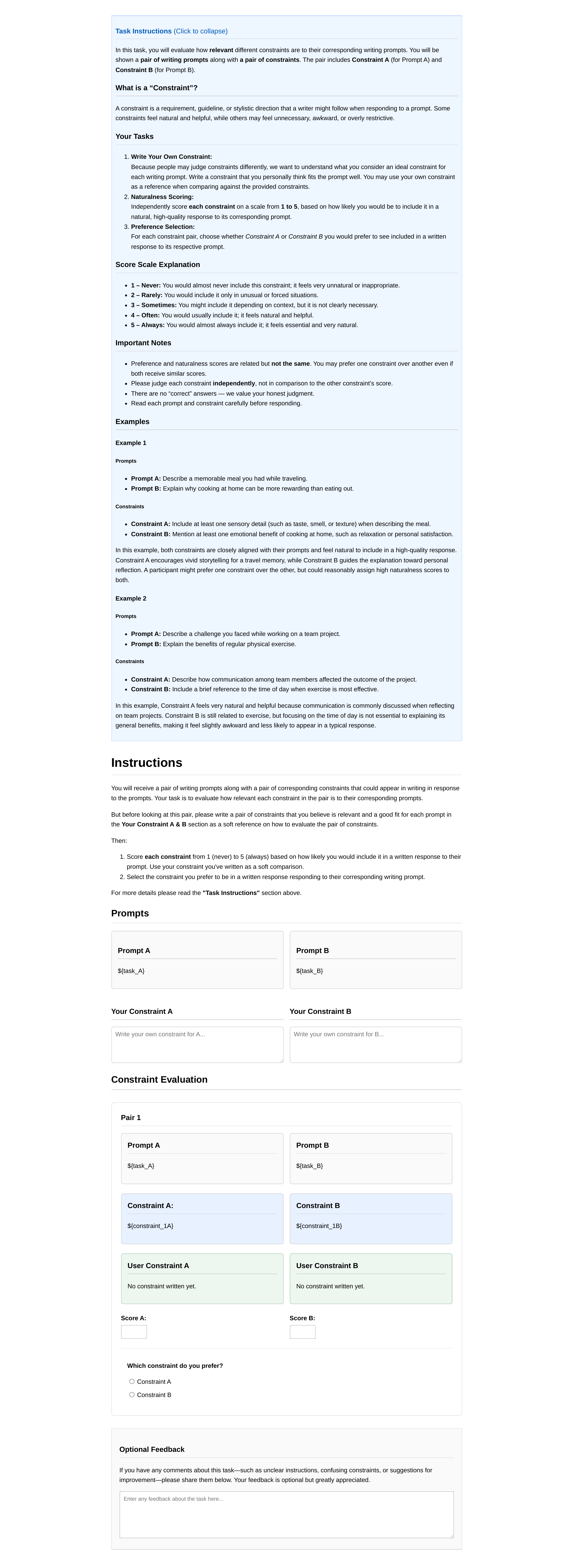}
    \caption{MTurk task template: user-written constraint reference.}
    \label{fig:user_constraints}
\end{figure}

\begin{figure}[htbp]
    \centering
    \includegraphics[width=\linewidth]{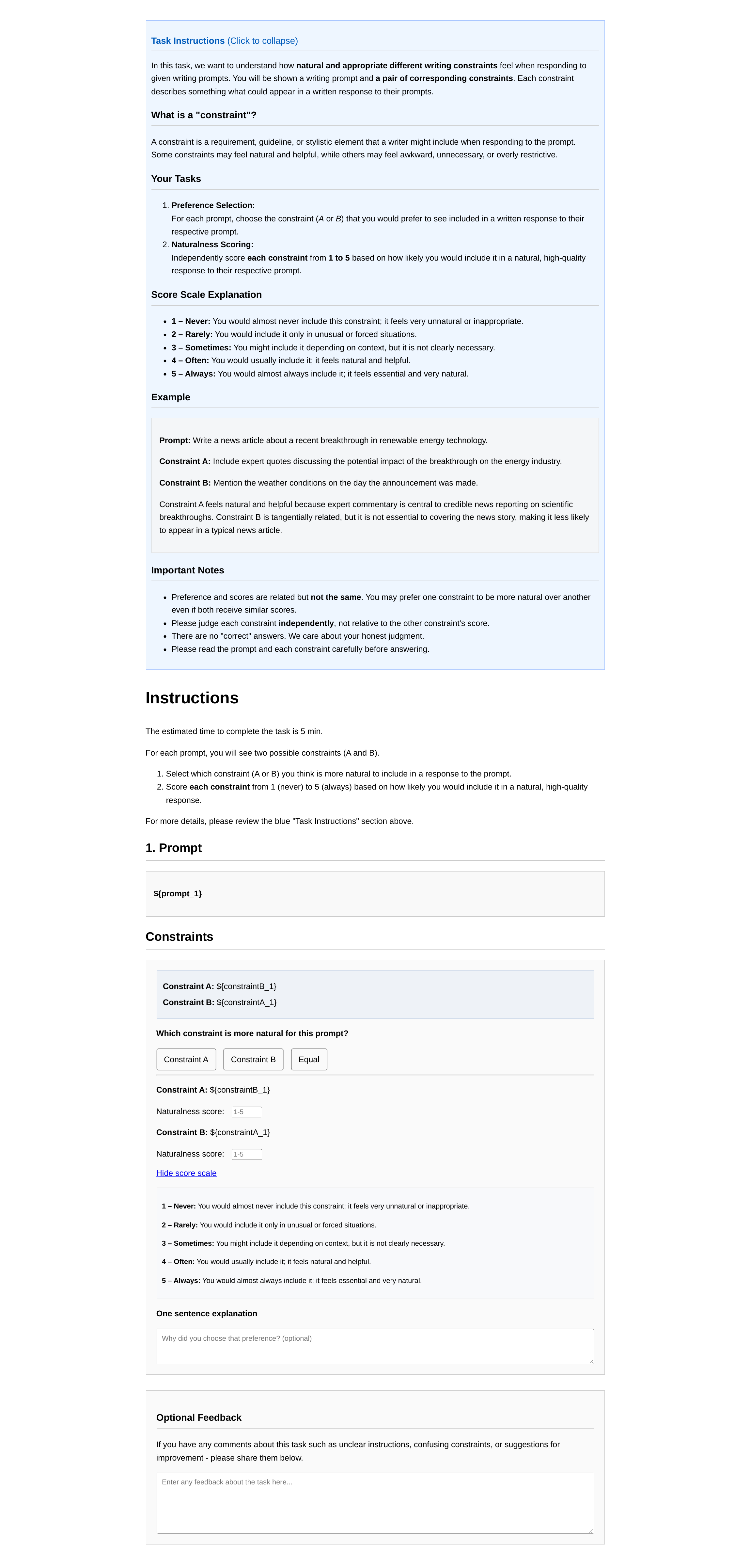}
    \caption{MTurk task template: naturalness \& preference scoring.}
    \label{fig:human_exp}
\end{figure}

We used two MTurk task templates for our naturalness evaluation, shown in Figures~\ref{fig:user_constraints} and~\ref{fig:human_exp}.

\textbf{Collecting human-written constraints.} To obtain a human reference point, we first ran a collection batch in which workers were shown two writing prompts side by side and asked to write their own constraint for each prompt before evaluating a provided constraint pair. The first template (Figure~\ref{fig:user_constraints}) was used for the collection batch. Prompt A corresponded to the main task under the Single condition and Prompt B corresponded to the main task under the Common condition, so that the human-written constraint for each prompt could later be compared against constraints synthesized under the matching condition. Copy-paste was disabled so that workers produced genuine, independently written constraints rather than echoing the prompt or a shown constraint. Across the $25$ tasks (HITs) in this batch, this yielded $50$ human-written constraints, one for each of Prompt A and Prompt B per HIT. This step served as a soft reference point for the worker's own judgment. To reduce noise, we only allow master workers to do the tasks and pay \$1 for each HIT to make the hourly salary close to \$12.

\textbf{Preference collection.} In the main annotation batch, each HIT presented a single writing prompt paired with two candidate constraints, a human-written constraint and a constraint synthesized by one of our conditions (\textbf{Single+None}, \textbf{Common+None}, \textbf{Common+Eval}, \textbf{Common+Direct}), and workers judged which felt more natural for that prompt, or selected Equal if both fit comparably well.

%Table~\ref{tb:human_exp} reports the resulting Human Win, Tie, and LLM Win rates for each condition.
The second template (Figure~\ref{fig:human_exp}) was used for the main preference collection batch. Each task showed a single writing prompt paired with two candidate constraints, A and B. Workers selected which constraint felt more natural for that prompt, with an Equal option available if neither was clearly better, and independently scored each constraint on the same 1 to 5 scale. An optional one sentence explanation of the worker's preference was also collected.
While the example in Figure~\ref{fig:human_exp} shows a single prompt for illustration, each HIT in this template presented four such prompt sections, each with its own constraint pair and preference selection.
Both templates included worked examples in the instructions to illustrate how naturalness should be judged, and both randomized the display order of Constraint A and Constraint B to prevent position bias.

\section{AI Usage}
We use Claude code to generate some analysis codes and MTurk Template. We also use Claude and ChatGPT to help us search for some related work, or provide writing suggestions.

\end{document}